\documentclass[10pt,journal,cspaper,compsoc]{IEEEtran}
\ifCLASSOPTIONcompsoc
  \usepackage[nocompress]{cite}
\else
\fi
\ifCLASSINFOpdf
    \usepackage[pdftex]{graphicx}
    \DeclareGraphicsExtensions{.pdf,.jpeg,.png}
\else
    \usepackage[dvips]{graphicx}
    \DeclareGraphicsExtensions{.eps}
\fi
\usepackage[cmex10]{amsmath}
\usepackage{algorithmic}
\usepackage{url}
\usepackage{multirow}
\usepackage{algorithm}

\usepackage{color}
\usepackage{amsmath}
\usepackage{amssymb}
\usepackage{booktabs}
\usepackage{comment}
\allowdisplaybreaks

\begin{document}
\title{Interpretable CNNs for Object Classification}


\author{Quanshi Zhang, Xin Wang, Ying Nian Wu, Huilin Zhou, and Song-Chun Zhu, \textit{Fellow, IEEE}
\IEEEcompsocitemizethanks{\IEEEcompsocthanksitem Quanshi Zhang, Xin Wang, and Huilin Zhou is with the John Hopcroft Center and the MoE Key Lab of Artificial Intelligence, AI Institute, at the Shanghai Jiao Tong University, China. Ying Nian Wu and Song-Chun Zhu are with the University of California, Los Angeles, USA.}
}% <-this % stops a space


\markboth{IEEE TRANSACTIONS ON PATTERN ANALYSIS AND MACHINE INTELLIGENCE}%
{Shell \MakeLowercase{\textit{et al.}}: Bare Demo of IEEEtran.cls for Computer Society Journals}

%\IEEEpubid{\makebox[\columnwidth]{\hfill 0000--0000/00/\$00.00~\copyright~2007 IEEE}%

% PRIOR to the title within the \IEEEcompsoctitleabstractindextext IEEEtran
\IEEEcompsoctitleabstractindextext{%

\begin{abstract}
This paper proposes a generic method to learn interpretable convolutional filters in a deep convolutional neural network (CNN) for object classification, where each interpretable filter encodes features of a specific object part. Our method does not require additional annotations of object parts or textures for supervision. Instead, we use the same training data as traditional CNNs. Our method automatically assigns each interpretable filter in a high conv-layer with an object part of a certain category during the learning process. Such explicit knowledge representations in conv-layers of CNN help people clarify the logic encoded in the CNN, \emph{i.e.} answering what patterns the CNN extracts from an input image and uses for prediction. We have tested our method using different benchmark CNNs with various structures to demonstrate the broad applicability of our method. Experiments have shown that our interpretable filters are much more semantically meaningful than traditional filters.
\end{abstract}

% not want either math or citations to appear in the abstract.

% Note that keywords are not normally used for peer review papers.
}

\maketitle
\IEEEdisplaynotcompsoctitleabstractindextext
\IEEEpeerreviewmaketitle

\section{Introduction}

In recent years, convolutional neural networks (CNNs)~\cite{CNN,CNNImageNet,ResNet} have achieved superior performance in many visual tasks, such as object classification and detection. In spite of the good performance, a deep CNN has been considered a black-box model with weak feature interpretability for decades. Boosting the feature interpretability of a deep model gradually attracts increasing attention recently, but it presents significant challenges for state-of-the-art algorithms.

In this paper, we focus on a new task, \emph{i.e.} without any additional annotations for supervision, revising a CNN to make its high conv-layers (\emph{e.g.} the top two conv-layers) encode interpretable object-part knowledge. The revised CNN is termed an \textit{interpretable CNN}.

More specifically, we propose a generic interpretable layer to ensure each filter in the proposed interpretable layer learns specific, discriminative object-part features. Filters in the interpretable layer are supposed to have some introspection of their feature representations and regularize their features towards object parts during the end-to-end learning. We trained interpretable CNNs on several benchmark datasets, and experimental results show that filters in the interpretable layer consistently represented the same object part across input images.

Note that our task of improving feature interpretability of a CNN is essentially different from the conventional visualization~\cite{CNNVisualization_1,CNNVisualization_2,CNNVisualization_3,FeaVisual,visualCNN_grad,visualCNN_grad_2} and diagnosis~\cite{Interpretability,CNNInfluence,trust,shap} of pre-trained CNNs. The interpretable CNN learns more interpretable features, whereas previous methods mainly explain pre-trained neural networks, instead of learning new and more interpretable features.

\begin{figure}[t]
\centering
\includegraphics[width=\linewidth]{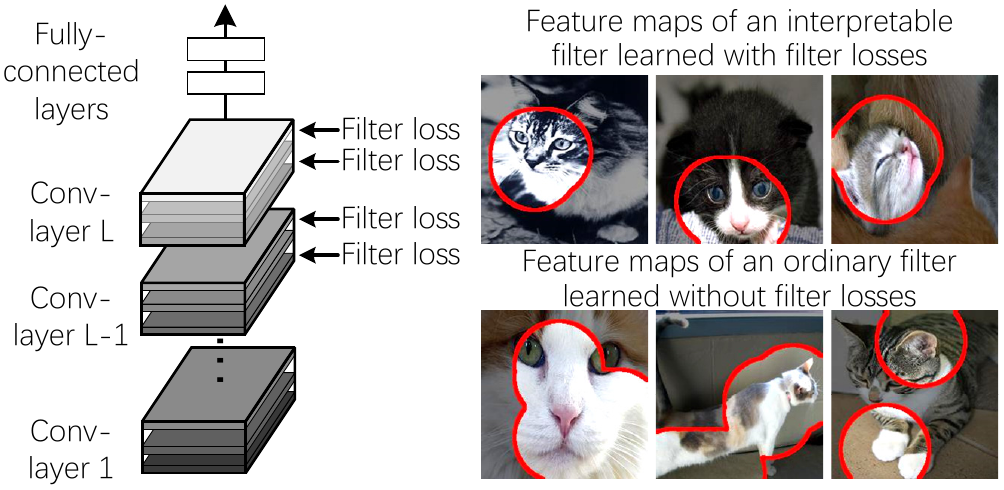}
\caption{Comparison of an interpretable filter's feature maps with a filter's feature maps in a traditional CNN.}
\label{fig:top}
\end{figure}

In addition, as discussed in \cite{Interpretability}, filters in low conv-layers usually describe textural patterns, while filters in high conv-layers are more likely to represent part patterns. Therefore, we focus on part-based interpretability and propose a method to ensure each filter in a high conv-layer to represent an object part.

Fig.~\ref{fig:top} visualizes the difference between a traditional filter and our interpretable filter. In a traditional CNN, a filter usually describes a mixture of patterns. For example, the filter may be activated by both the head part and the leg part of a cat. In contrast, the filter in our interpretable CNN is expected to be activated by a certain part.

Thus, the goal of this study can be summarized as follows. We propose a generic interpretable conv-layer to construct the interpretable CNN. Feature representations of the interpretable conv-layers are interpretable, \emph{i.e.} each filter in the interpretable conv-layer learns to consistently represent the same object part across different images. In addition, the interpretable conv-layer needs to satisfy the following properties:
\begin{itemize}
\item The interpretable CNN needs to be learned without any additional annotations of object parts for supervision. We use the same training samples as the original CNN for learning.
\item The interpretable CNN does not change the loss function of the classification task, and it can be broadly applied to different benchmark CNNs with various structures.
\item As an exploratory research, learning strict representations of object parts may hurt a bit the discrimination power. However, we need to control the decrease within a small range.
\end{itemize}

\begin{figure}[t]
\centering
\includegraphics[width=\linewidth]{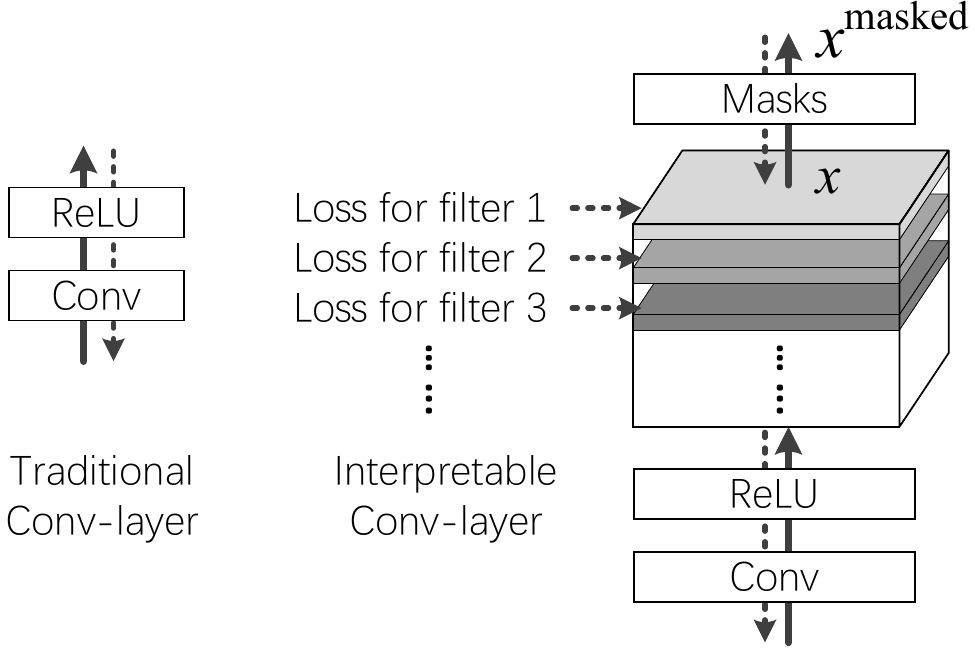}
\caption{Structures of an ordinary conv-layer and an interpretable conv-layer. Solid and dashed lines indicate the forward and backward propagations, respectively. During the forward propagation, our CNN assigns each interpretable filter with a specific mask \emph{w.r.t.} each input image during the learning process.}
\label{fig:net}
\end{figure}

\textbf{Method:} As shown in Fig.~\ref{fig:net}, we propose a simple yet effective loss. We simply add the loss to the feature map of each filter in a high conv-layer, so as to construct an interpretable conv-layer. The filter loss is proposed based the assumption that only a single target object is contained in the input image. The filter loss pushes the filter towards the representation of a specific object part.

Theoretically, we can prove that the loss encourages a low entropy of inter-category activations and a low entropy of spatial distributions of neural activations. In other words, this loss ensures that (i) each filter must encode an object part of a single object category, instead of representing multiple categories; (ii) The feature must consistently be triggered by a single specific part across multiple images, rather than be simultaneously triggered by different object regions in each input image. It is assumed that repetitive patterns on different object regions are more likely to describe low-level textures, instead of high-level parts.

\textbf{Value of feature interpretability:} Such explicit object-part representations in conv-layers of CNN can help people clarify the decision-making logic encoded in the CNN at the object-part level. Given an input image, the interpretable conv-layer enable people to explicitly identify which object parts are memorized and used by the CNN for classification without ambiguity. Note that the automatically learned object part may not have an explicit name, \emph{e.g.} a filter in the interpretable conv-layer may describe a partial region of a semantic part or the joint of two parts.

In critical applications, clear disentanglement of visual concepts in high conv-layers helps people trust a network's prediction. As analyzed in \cite{CNNBias}, a good performance on testing images cannot always ensure correct feature representations considering potential dataset bias. For example, in \cite{CNNBias}, a CNN used an unreliable context---eye features---to identify the ``lipstick'' attribute of a face image. Therefore, people need to semantically and visually explain what patterns are learned by the CNN.

\textbf{Contributions:} In this paper, we focus on a new task, \emph{i.e.} end-to-end learning an interpretable CNN without any part annotations, where filters of high conv-layers represent specific object parts. We propose a simple yet effective method to learn interpretable filters, and the method can be broadly applied to different benchmark CNNs. Experiments show that our approach has significantly boosted feature interpretability of CNNs.

A preliminary version of this paper appeared in \cite{interpretableCNN}.

\section{Related work}

The interpretability and the discrimination power are two crucial aspects of a CNN~\cite{Interpretability}. In recent years, different methods are developed to explore the semantics hidden inside a CNN. Our previous paper~\cite{InterpretabilitySurvey} provides a comprehensive survey of recent studies in exploring visual interpretability of neural networks, including (i) the visualization and diagnosis of CNN representations, (ii) approaches for disentangling CNN representations into graphs or trees, (iii) the learning of CNNs with disentangled and interpretable representations, and (iv) middle-to-end learning based on model interpretability.

\subsection{Interpretation of pre-trained neural networks}

\textbf{Network visualization:}{\verb| |} Visualization of filters in a CNN is the most direct way of exploring the pattern that is encoded by the filter. Gradient-based visualization~\cite{CNNVisualization_1,CNNVisualization_2,CNNVisualization_3} showed the appearance that maximized the score of a given unit. Furthermore, Bau \emph{et al.}~\cite{Interpretability} defined and analyzed the interpretability of each filter. They classified all potential semantics into the following six types, \textit{objects}, \textit{parts}, \textit{scenes}, \textit{textures}, \textit{materials}, and \textit{colors}. We can further summarize the semantics of \textit{objects} and \textit{parts} as part patterns with specific contours and consider the other four semantics as textural patterns without explicit shapes. Recently, \cite{olah2017feature} provided tools to visualize filters of a CNN. Dosovitskiy~\emph{et al.}~\cite{FeaVisual} proposed up-convolutional nets to invert feature maps of conv-layers to images. However, up-convolutional nets cannot mathematically ensure the visualization result reflects actual neural representations.

Although above studies can produce clear visualization results, theoretically, gradient-based visualization of a filter usually selectively visualizes the strongest activations of a filter in a high conv-layer, instead of illustrating knowledge hidden behind all activations of the filter; otherwise, the visualization result will be chaotic. Similarly, \cite{Interpretability} selectively analyzed the semantics of the highest 0.5\% activations of each filter. In comparisons, we aim to purify the semantic meaning of each filter in a high conv-layer, \emph{i.e.} letting most activations of a filter be explainable, instead of extracting meaningful neural activations for visualization.

\textbf{Pattern retrieval:}{\verb| |} Unlike passive visualization, some methods actively retrieve certain units with certain meanings from CNNs. Just like mid-level features~\cite{MiddleLevel} of images, pattern retrieval mainly focuses on mid-level representations in conv-layers. For example, Zhou~\emph{et al.}~\cite{CNNSemanticDeep,CNNSemanticDeep2} selected units from feature maps to describe ``scenes''. Simon~\emph{et al.} discovered objects from feature maps of conv-layers~\cite{ObjectDiscoveryCNN_2}, and selected certain filters to represent object parts~\cite{CNNSemanticPart}. Zhang~\emph{et al.}~\cite{CNNAoG} extracted certain neural activations of a filter to represent object parts in a weakly-supervised manner. They also disentangled CNN representations via active question-answering and summarized the disentangled knowledge using an And-Or graph~\cite{DeepQA}. \cite{holdingHands} used human interactions to refine the AOG representation of CNN knowledge. \cite{interpretVQA_grad} used a gradient-based method to explain visual question-answering. Other studies~\cite{explainableFeature,explainableFeature2,explainableFeature3,explainableFeature4} selected filters or neural activations with specific meanings from CNNs for various applications. Unlike the retrieval of meaningful neural activations from noisy features, our method aims to substantially boost the interpretability of features in intermediate conv-layers.

\textbf{Model diagnosis:}{\verb| |} Many approaches have been proposed to diagnose CNN features, including exploring semantic meanings of convolutional filters~\cite{CNNAnalysis_1}, evaluating the transferability of filters~\cite{CNNAnalysis_2}, and the analysis of feature distributions of different categories~\cite{CNNVisualization_5}. The LIME~\cite{trust} and the SHAP~\cite{shap} are general methods to extract input units of a neural network that are used for the prediction score. For CNNs oriented to visual tasks, gradient-based visualization methods~\cite{visualCNN_grad,visualCNN_grad_2} and \cite{ExplainingArea} extracted image regions that are responsible for the network output, in order to clarify the logic of network prediction. These methods require people to manually check image regions accountable for the label prediction for each testing image. \cite{CNNDiagnosis} extracted relationships between representations of various categories from a CNN. In contrast, given an interpretable CNN, people can directly identify object parts or filters that are used for prediction.

As discussed by Zhang~\emph{et al.}~\cite{CNNBias}, knowledge representations of a CNN may be significantly biased due to dataset bias, even though the CNN sometimes exhibits good performance. For example, a CNN may extract unreliable contextual features for prediction. Network-attack methods~\cite{pixelAttack,CNNInfluence,CNNAnalysis_1} diagnosed network representation flaws using adversarial samples of a CNN. For example, influence functions~\cite{CNNInfluence} can be used to generate adversarial samples, in order to fix the training set and further debug representations of a CNN. \cite{banditUnknown} discovered blind spots of knowledge representation of a pre-trained CNN in a weakly-supervised manner.

\textbf{Distilling neural networks into explainable models:}{\verb| |} Furthermore, some method distilled CNN knowledge into another model with interpretable features for explanations. \cite{additiveExplainer} distilled knowledge of a neural network into an additive model to explain the knowledge inside the network. \cite{explanatoryTree} roughly represented the rationale of each CNN prediction using a semantic tree structure. Each node in the tree represented a decision-making mode of the CNN. Similarly, \cite{explanatoryGraph} used a semantic graph to summarize and explain all part knowledge hidden inside conv-layers of a CNN.

\subsection{Learning interpretable feature representations}

Unlike the diagnosis and visualization of pre-trained CNNs, some approaches were developed to learn meaningful feature representations in recent years. Automatically learning interpretable feature representations without additional human annotations proposes new challenges to state-of-the-art algorithms. For example, \cite{rightReason} required people to label dimensions of the input that were related to each output, in order to learn a better model. Hu~\emph{et al.}~\cite{LogicRuleNetwork} designed logic rules to regularize network outputs during the learning process. Sabour~\emph{et al.}~\cite{capsule} proposed a capsule model, where each feature dimension of a capsule may represent a specific meaning. Similarly, we invent a generic filter loss to regularize the representation of a filter to improve its interpretability.

In addition, unlike the visualization methods (\emph{e.g.} the Grad-CAM method~\cite{visualCNN_grad_2}) using a single saliency map for visualization, our interpretable CNN disentangles feature representations and uses different filters to represent the different object parts.

\section{Algorithm}

Given a target conv-layer of a CNN, we expect each filter in the conv-layer to be activated by a certain object part of a certain category, while remain inactivated on images of other categories\footnote[1]{To avoid ambiguity, we evaluate or visualize the semantic meaning of each filter by using the feature map after the ReLU and mask operations.}. Let {${\bf I}$} denote a set of training images, where {${\bf I}_{c}\subset{\bf I}$} represents the subset that belongs to category $c$, ({$c=1,2,\ldots,C$}). Theoretically, we can use different types of losses to learn CNNs for multi-class classification and binary classification of a single class (\emph{i.e.} {$c=1$} for images of a category and {$c=2$} for random images).

In the following paragraphs, we focus on the learning of a single filter $f$ in a conv-layer. Fig.~\ref{fig:net} shows the structure of our interpretable conv-layer. We add a loss to the feature map $x$ of the filter $f$ after the ReLU operation. The filter loss $Loss_{f}$ pushes the filter $f$ to represent a specific object part of the category $c$ and keep silent on images of other categories. Please see Section~\ref{sec:learning} for the determination of the category $c$ for the filter $f$. Let ${\bf X}=\{x|x=f(I)\in\mathbb{R}^{n\times n},I\in{\bf I}\}$ denote a set of feature maps of $f$ after an ReLU operation \emph{w.r.t.} different images. Given an input image $I\in{\bf I}_{c}$, the feature map in an intermediate layer $x=f(I)$ is an $n\times n$ matrix, $x_{ij}\geq0$. If the target part appears, we expect the feature map $x=f(I)$ to exclusively activate at the target part's location; otherwise, the feature map should keep inactivated.

Therefore, a high interpretability of the filter $f$ requires a high mutual information between the feature map $x=f(I)$ and the part location, \emph{i.e.} the part location can roughly determine activations on the feature map $x$.

Accordingly, we formulate the filter loss as the minus mutual information, as follows.
\begin{equation}
\!\!\!{\bf Loss}_{f}\!=\!-MI({\bf X};{\boldsymbol\Omega})\!=\!-\sum_{\mu\in{\boldsymbol\Omega}}p(\mu)\sum_{x}p(x|\mu)\log\frac{p(x|\mu)}{p(x)}\!
\label{eqn:loss}
\end{equation}
where $MI(\cdot)$ denotes the mutual information; ${\boldsymbol\Omega}=\{\mu_1,\mu_2,\ldots,\mu_{n^2}\}\cup\{\mu^{-}\}$. We use $\mu_1,\mu_2,\ldots,\mu_{n^2}$ to denote the $n^2$ neural units on the feature map $x$, each $\mu=[i,j]\in{\boldsymbol\Omega}$, $1\leq i,j\leq n$, corresponding to a location candidate for the target part. $\mu^{-}$ denotes a dummy location for the case when the target part does not appear on the image.

Given an input image, the above loss forces each filter to match and only match one of the templates, \emph{i.e.} making the feature map of the filter contain a single significant activation peak at most. This ensures each filter to represent a specific object part.

${\boldsymbol\bullet}$ $p(\mu)$ measures the probability of the target part appearing at the location $\mu$. If annotations of part locations are given, then the computation of $p(\mu)$ is simple. People can manually assign a semantic part with the filter $f$, and then $p(\mu)$ can be determined using part annotations.

However, in our study, the target part of filter $f$ is not pre-defined before the learning process. Instead, the part corresponding to $f$ needs to be determined during the learning process. More crucially, we do not have any ground-truth annotations of the target part, which boosts the difficulty of calculating $p(\mu)$.

\begin{figure}[t]
\centering
\includegraphics[width=\linewidth]{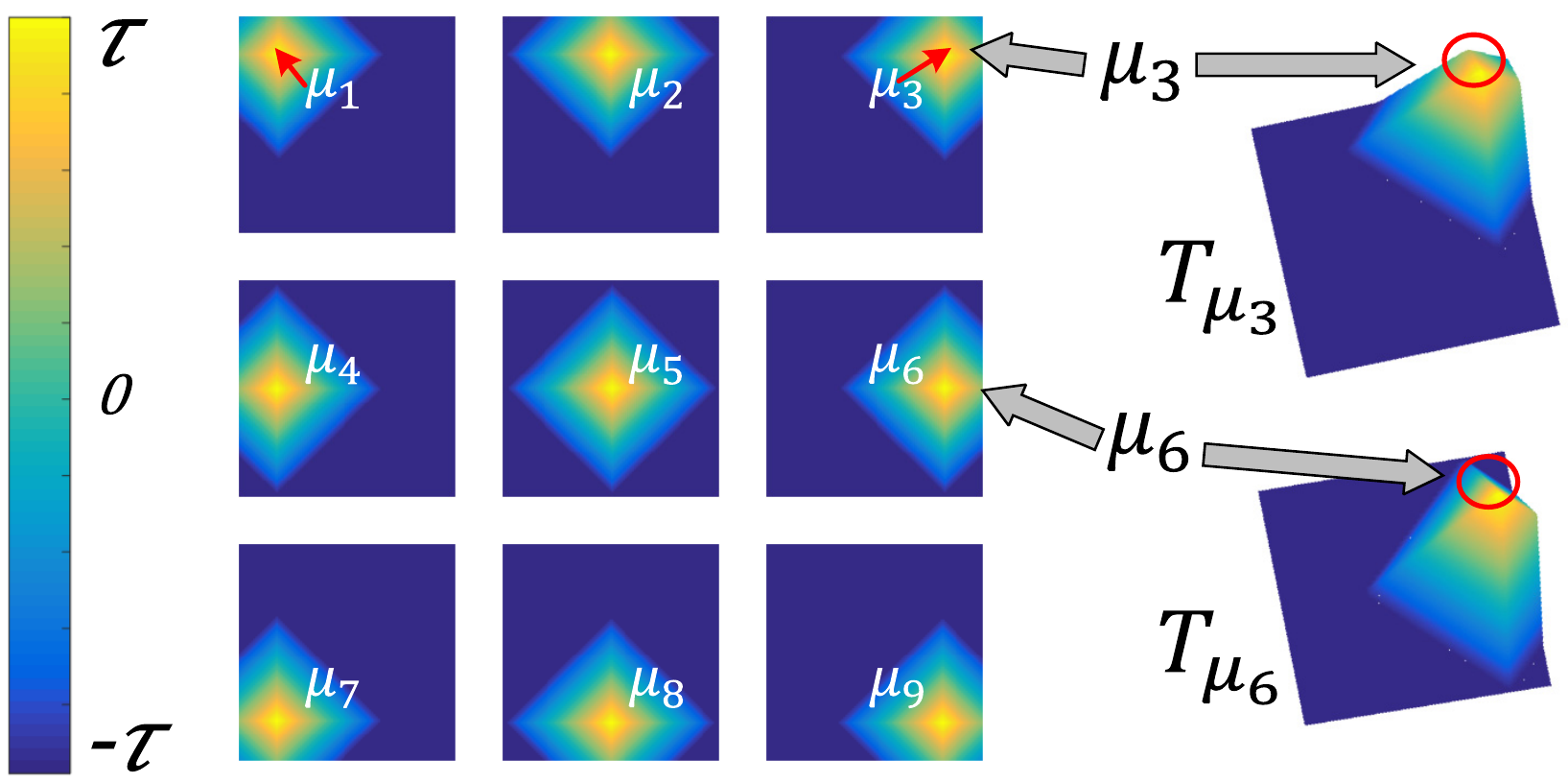}
\caption{Templates of {$T_{\mu_{i}}$}. We show a toy example of $n=3$. Each template {$T_{\mu_{i}}$} matches to a feature map $x$ when the target part mainly triggers the $i$-th unit in $x$. In fact, the algorithm also supports a round template based on the L-2 norm distance. Here, we use the L-1 norm distance instead to speed up the computation.}
\label{fig:template}
\end{figure}

${\boldsymbol\bullet}$ The conditional likelihood $p(x|\mu)$ measures the fitness between a feature map $x$ and the part location $\mu\in{\boldsymbol\Omega}$. In order to simplify the computation of $p(x|\mu)$, we design $n^2$ templates for $f$, $\{T_{\mu_1},T_{\mu_2},\ldots,T_{\mu_{n^2}}\}$. As shown in Fig.~\ref{fig:template}, each template $T_{\mu_{i}}$ is an $n\times n$ matrix. $T_{\mu_{i}}$ describes the ideal distribution of activations for the feature map $x$ when the target part mainly triggers the $i$-th unit in $x$. In addition, we also design a negative template $T^{-}$ corresponding to the dummy location $\mu^{-}$. The feature map can match to $T^{-}$, when the target part does not appear on the input image. In this study, the prior probability is given as {\small$p(\mu_{i})\!=\!\frac{\alpha}{n^2}, p(\mu^{-})\!=\!1-\alpha$}, where $\alpha$ is a constant prior likelihood.

Note that in Equation~\eqref{eqn:loss}, we do not manually assign filters with different categories. Instead, we use the negative template $\mu^{-}$ to help the assignment of filters. \emph{I.e.} the negative template ensures that each filter represents a specific object part (if the input image does not belong to the target part, then the input image is supposed to match $\mu^{-}$), which also ensures a clear assignment of filters to categories. Here, we assume two categories do not share object parts, \emph{e.g.} eyes of dogs and those of cats do not have similar contextual appearance.

We define $p(x|\mu)$ below, which follows a standard form widely used in \cite{MiningAOG,InhomogeneousFRAME}.
\begin{equation}
p(x|\mu)\approx p(x|T_{\mu})=\frac{1}{Z_{\mu}}\exp\big[tr(x\cdot T_{\mu})\big]
\end{equation}
where $Z_{\mu}=\sum_{x\in{\bf X}}\exp[tr(x\cdot T_{\mu})]$. $tr(\cdot)$ indicates the trace of a matrix, and $tr(x\cdot T_{\mu})=\sum_{ij}x_{ij}t_{ji}$, $x,T_{\mu}\in\mathbb{R}^{n\times n}$. $p(x)=\sum_{\mu}p(\mu)p(x|\mu)$.

\textbf{Part templates:} As shown in Fig.~\ref{fig:template}, a negative template is given as $T^{-}=(t^{-}_{ij})$, $t^{-}_{ij}=-\tau<0$, where $\tau$ is a positive constant. A positive template corresponding to $\mu$ is given as {$T_{\mu}=(t^{+}_{ij})$}, $t^{+}_{ij}=\tau\cdot\max(1-\beta\frac{\Vert[i,j]-\mu\Vert_1}{n},-1)$, where $\Vert\cdot\Vert_1$ denotes the L-1 norm distance. Note that the lowest value in a positive template is -1 instead of 0. It is because that the negative value in the template penalizes neural activations outside the domain of the highest activation peak, which ensures each filter mainly has at most a single significant activation peak.

\subsection{Part localization \& the mask layer}

Given an input image $I$, the filter $f$ computes a feature map $x$ after the ReLU operation. Without ground-truth annotations of the target part for $f$, in this study, we determine the part location on $x$ during the learning process. We consider the neural unit with the strongest activation $\hat{\mu}={\arg\!\max}_{\mu=[i,j]}x_{ij}$, $1\leq i,j\leq n$ as the target part location.

As shown in Fig.~\ref{fig:net}, we add a mask layer above the interpretable conv-layer. Based on the estimated part position $\hat{\mu}$, the mask layer assigns a specific mask with $x$ to filter out noisy activations. The mask operation is separate from the filter loss in Equation~\eqref{eqn:loss}. Our method selects the template $T_{\hat{\mu}}$ \emph{w.r.t.} the part location $\hat{\mu}$ as the mask. We compute $x^{\textrm{masked}}=\max\{x\circ T_{\hat{\mu}},0\}$ as the output masked feature map, where $\circ$ denotes the Hadamard (element-wise) product. The mask operation supports gradient back-propagations.

\begin{figure*}[t]
\centering
\includegraphics[width=0.8\linewidth]{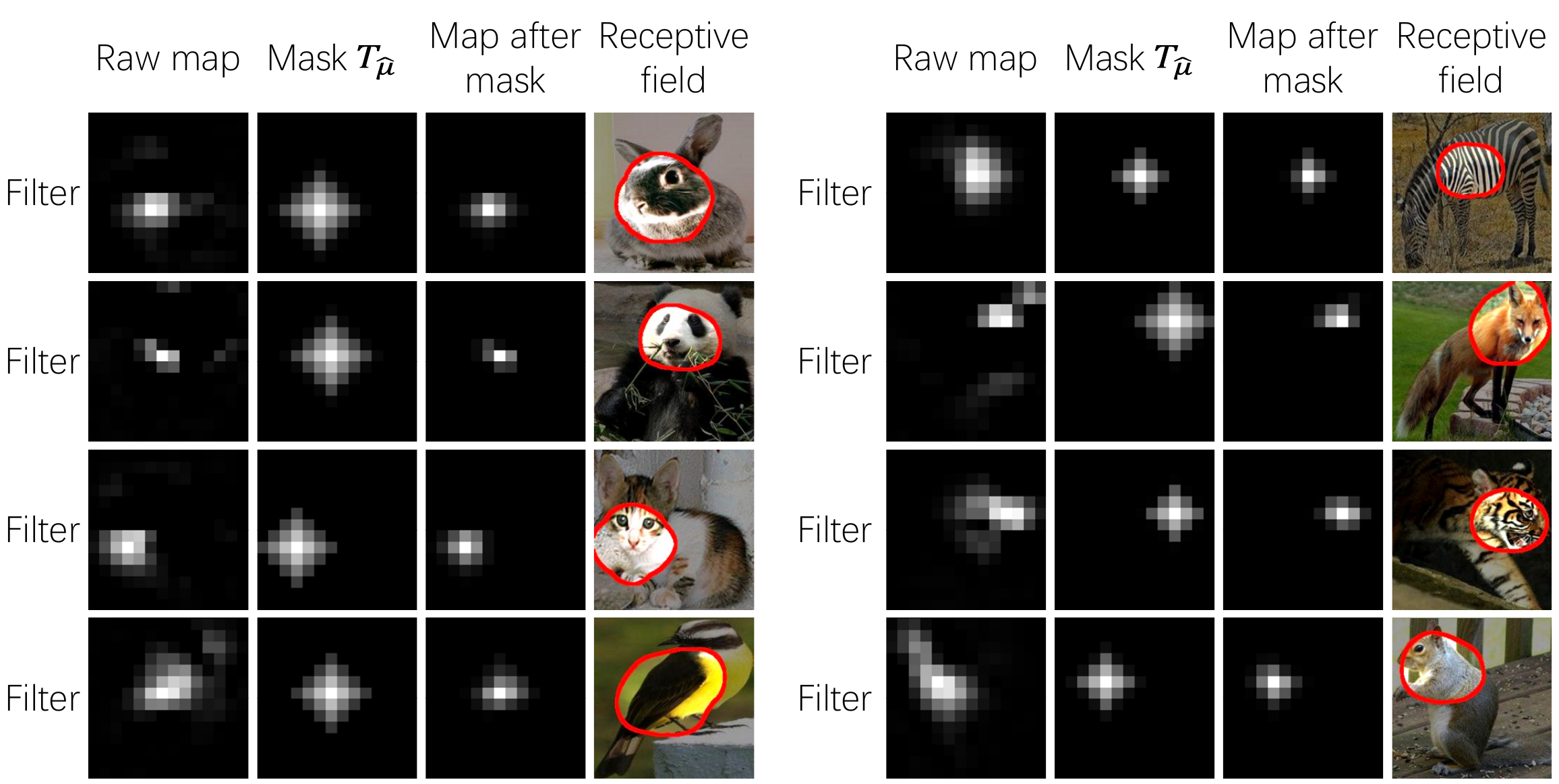}
\caption{Given an input image $I$, from left to right, we consequently show the feature map of a filter after the ReLU layer $x$, the assigned mask {$T_{\hat{\mu}}$}, the masked feature map {$x^{\textrm{masked}}$}, and the image-resolution RF of activations in {$x^{\textrm{masked}}$} computed by \cite{CNNSemanticDeep}.}
\label{fig:map}
\end{figure*}

Fig.~\ref{fig:map} visualizes the masks $T_{\hat{\mu}}$ chosen for different images, and compares the original and masked feature maps. The CNN selects different templates for different images.

Note that although a filter usually has much stronger neural activations on the target category than on other categories, the magnitude of neural activations is still not discriminative enough for classification. Moreover, during the testing process, people do not have ground-truth class labels of input images. Thus, to ensure stable feature extraction, our method only selects masks from the $n^2$ positive templates $\{T_{\mu_{i}}\}$ and omits the negative template $T^{-}$ for all images, no matter whether or not input images contain the target part. Such an operation is conducted during the forward process for both training and testing processes.

\subsection{Learning}
\label{sec:learning}

We train the interpretable CNN in an end-to-end manner. During the forward-propagation process, each filter in the CNN passes its information in a bottom-up manner, just like traditional CNNs. During the back-propagation, each filter in an interpretable conv-layer receives gradients \emph{w.r.t.} its feature map $x$ from both the final task loss ${\bf L}(\hat{y}_{k},y^{*}_{k})$ on the $k$-th sample and the filter loss, ${\bf Loss}_{f}$, as follows:
\begin{equation}
\frac{\partial{\bf Loss}}{\partial x_{ij}}=\lambda\sum_{f}\frac{\partial{\bf Loss}_{f}}{\partial x_{ij}}+\frac{1}{N}\sum_{k=1}^{N}\frac{\partial{\bf L}(\hat{y}_{k},y^{*}_{k})}{\partial x_{ij}}
\end{equation}
where $\lambda$ is a weight. Then, we back propagate $\frac{\partial{\bf Loss}}{\partial x_{ij}}$ to lower layers and compute gradients \emph{w.r.t.} feature maps and gradients \emph{w.r.t.} parameters in lower layers to update the CNN.

For implementation, gradients of ${\bf Loss}_{f}$ \emph{w.r.t.} each element $x_{ij}$ of feature map $x$ are computed as follows.
\begin{small}
\begin{eqnarray}
\!\!\frac{\partial{\bf Loss}_{f}}{\partial x_{ij}}\!=\!\frac{1}{Z_{\mu}}\!\sum_{\mu}p(\mu)t_{ij}e^{tr(x\cdot T_{\mu})}\!\Big\{tr(x\cdot T_{\mu})\!-\!\log\big[Z_{\mu}p(x)\big]\!\Big\}\!\!\!\!\!\!\!\!\!\nonumber\\
\approx\frac{p(\hat{\mu})\hat{t}_{ij}}{Z_{\hat{\mu}}}e^{tr(x\cdot T_{\hat{\mu}})}\Big\{tr(x\cdot T_{\hat{\mu}})-\log{Z_{\hat{\mu}}}-\log{p(x)}\Big\}
\label{eqn:grad}
\end{eqnarray}
\end{small}
where $T_{\hat{\mu}}$ is the target template for feature map $x$. If the input image $I$ belongs to the target category of filter $f$, then $\hat{\mu}={\arg\!\max}_{\mu=[i,j]}x_{ij}$. If image $I$ belongs to other categories, then $\hat{\mu}=\mu^{-}$. Please see the appendix for the proof of the above equation.

Considering $\forall \mu\in{\boldsymbol\Omega}\setminus\{\hat{\mu}\}$, $e^{tr(x\cdot T_{\hat{\mu}})}\gg e^{tr(x\cdot T_{\mu})}$ and $p(\hat{\mu})\gg p(\mu)$ after initial learning episodes, we make the above approximation to simplify the computation. Because $Z_{\hat{\mu}}$ is computed using numerous feature maps, we can roughly treat $Z_{\hat{\mu}}$ as a constant to compute gradients in the above equation. We gradually update the value of $Z_{\hat{\mu}}$ during the training process. More specifically, we can use a subset of feature maps to approximate the value of $Z_{\mu}$, and continue to update $Z_{\mu}$ when we receive more feature maps during the training process. Similarly, we can approximate $p(x)$ using a subset of feature maps. We can also approximate $p(x)=\sum_{\mu}p(\mu)p(x|\mu)=\sum_{\mu}p(\mu)\frac{\exp[tr(x\cdot T_{\mu})]}{Z_{\mu}}\approx\sum_{\mu}p(\mu)\mathbb{E}_{x}\frac{\exp[tr(x\cdot T_{\mu})]}{Z_{\mu}}$ without huge computation.

\textbf{Determining the target category for each filter:} We need to assign each filter $f$ with a target category $\hat{c}$ to approximate gradients in Equation~(\ref{eqn:grad}). We simply assign the filter $f$ with the category $\hat{c}$ whose images activate $f$ the most, \emph{i.e.} {$\hat{c}={\arg\!\max}_{c}\mathbb{E}_{x=f(I):I\in{\bf I}_{c}}\sum_{ij}x_{ij}$}.

\subsection{Understanding the filter loss}

The filter loss in Equation~(\ref{eqn:loss}) can be re-written as
\begin{equation}
\!\!\!\!{\bf Loss}_{f}\!=\!-H({\boldsymbol\Omega})\!+\!H({\boldsymbol\Omega}'|{\bf X})\!+\!\sum_{x}p({\boldsymbol\Omega}^{+}\!,x)H({\boldsymbol\Omega}^{+}|X\!=\!x)\!
\label{eqn:understand}
\end{equation}
where {${\boldsymbol\Omega}'=\{\mu^{-},{\boldsymbol\Omega}^{+}\}$}. {$H({\boldsymbol\Omega})=-\sum_{\mu\in{\boldsymbol\Omega}}p(\mu)\log p(\mu)$} is a constant prior entropy of part locations. Thus, the filter loss minimizes two conditional entropies, $H({\boldsymbol\Omega}'|{\bf X})$ and $H({\boldsymbol\Omega}^{+}|X=x)$. Please see the appendix for the proof of the above equation.

\noindent
${\boldsymbol\bullet}$\textbf{Low inter-category entropy:} The second term {$H({\boldsymbol\Omega}'=\{\mu^{-},{\boldsymbol\Omega}^{+}\}|{\bf X})$} is computed as
\begin{equation}
\!\!\!H({\boldsymbol\Omega}'\!=\!\{\mu^{-}\!,\!{\boldsymbol\Omega}^{+}\}|{\bf X})=-\sum_{x}p(x)\!\!\!\!\!\!\!\!\!\sum_{\mu\in\{\mu^{-},{\boldsymbol\Omega}^{+}\}}\!\!\!\!\!\!\!\!\!p(\mu|x)\log p(\mu|x)\!
\end{equation}
where ${\boldsymbol\Omega}^{+}=\{\mu_{1},\ldots,\mu_{n^2}\}\subset{\boldsymbol\Omega}$, $p({\boldsymbol\Omega}^{+}|x)=\sum_{\mu\in{\boldsymbol\Omega}^{+}}p(\mu|x)$. We define the set of all real locations ${\boldsymbol\Omega}^{+}$ as a single label to represent category $c$. We use the dummy location $\mu^{-}$ to roughly indicate matches to other categories.

This term encourages a low conditional entropy of inter-category activations, \emph{i.e.} a well-learned filter $f$ needs to be exclusively activated by a certain category $c$ and keep silent on other categories. We can use a feature map $x$ of $f$ to identify whether or not the input image belongs to category $c$, \emph{i.e.} $x$ fitting to either $T_{\hat{\mu}}$ or $T^{-}$, without significant uncertainty.

\noindent
${\boldsymbol\bullet}$\textbf{Low spatial entropy:} The third term in Equation~(\ref{eqn:understand}) is given as
\begin{equation}
H({\boldsymbol\Omega}^{+}|X\!=\!x)=\sum_{\mu\in{\boldsymbol\Omega}^{+}}\tilde{p}(\mu|x)\log\tilde{p}(\mu|x)
\end{equation}
where {$\tilde{p}(\mu|x)=\frac{p(\mu|x)}{p({\boldsymbol\Omega}^{+}|x)}$}. This term encourages a low conditional entropy of the spatial distribution of $x$'s activations. \emph{I.e.} given an image {$I\in{\bf I}_{c}$}, a well-learned filter should only be activated in a single region $\hat{\mu}$ of the feature map $x$, instead of being repetitively triggered at different locations.

\begin{figure*}[t]
\centering
\includegraphics[width=0.9\linewidth]{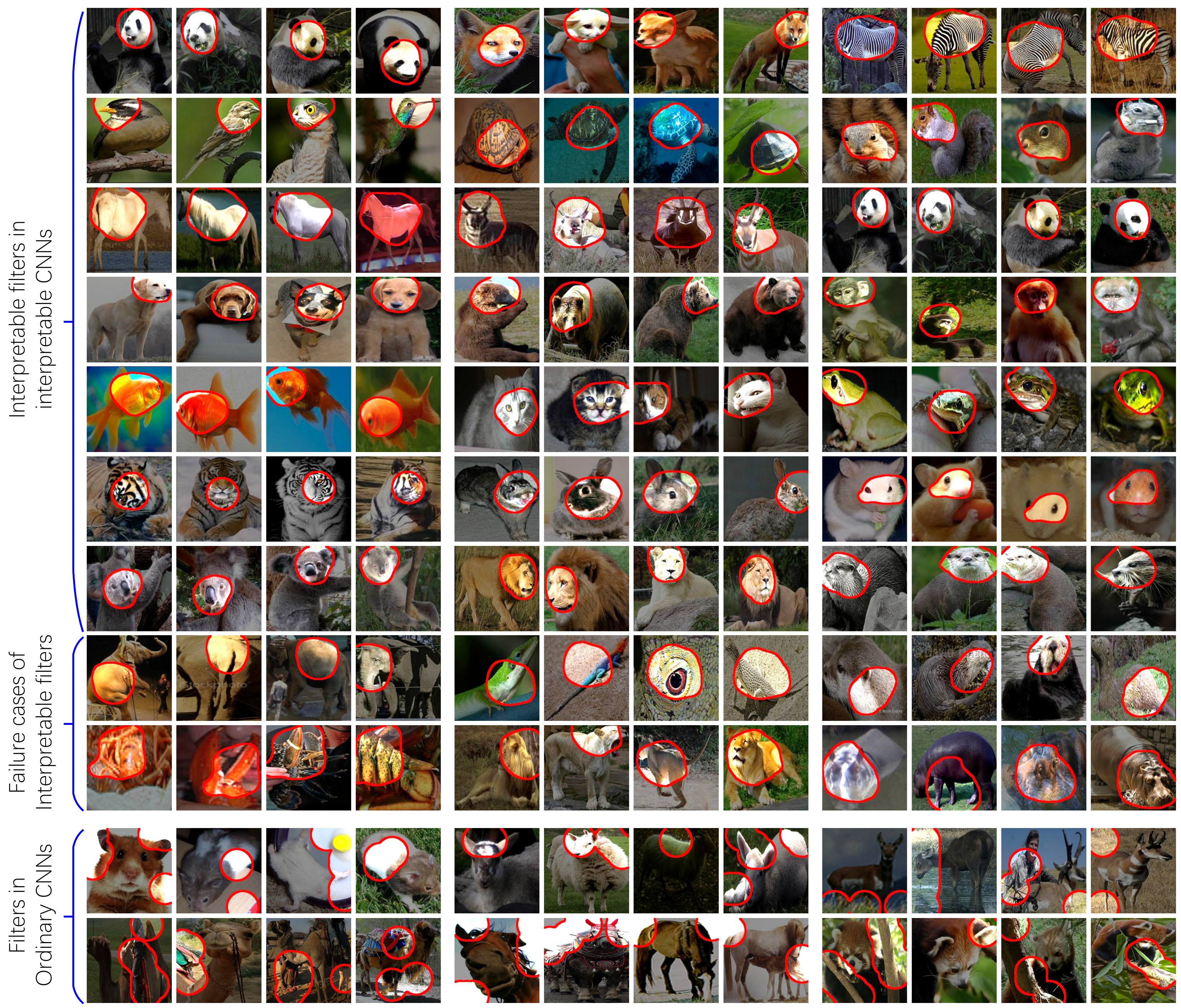}
\includegraphics[width=0.9\linewidth]{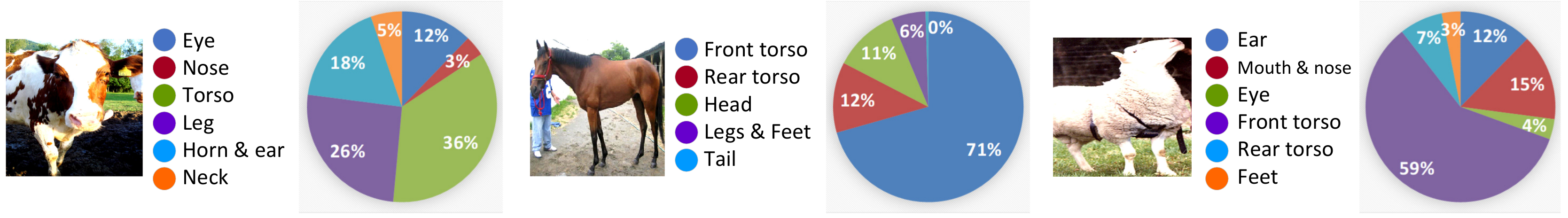}
\caption{Visualization of filters in top conv-layers (top) and quantitative contribution of object parts to the prediction (bottom). (top) We used \cite{CNNSemanticDeep} to estimate the image-resolution receptive field of activations in a feature map to visualize a filter's semantics. Each group of four feature maps for a category are computed using the same interpretable filter. These images show that each interpretable filter is consistently activated by the same object part through different images. Four rows visualize filters in interpretable CNNs, and two rows correspond to filters in ordinary CNNs. (bottom) The clear disentanglement of object-part representations help people to quantify the contribution of different object parts to the network prediction. We show the explanation for part contribution, which was generated by the method of \cite{SemanticalQuantitative}.}
\label{fig:visual}
\end{figure*}

\section{Experiments}

In experiments, we applied our method to modify four types of CNNs with various structures into interpretable CNNs and learned interpretable CNNs based on three benchmark datasets, in order to demonstrate the broad applicability. We learned interpretable CNNs for binary classification of a single category and multi-category classification. We used different techniques to visualize the knowledge encoded in interpretable filters, in order to qualitatively illustrate semantic meanings of these filters. Furthermore, we used two types of evaluation metrics, \emph{i.e.} the object-part interpretability and the location instability, to measure the clarity of the meaning of a filter.

Our experiments showed that an interpretable filter in our interpretable CNN usually consistently represented the same part through different input images, while a filter in an ordinary CNN mainly described a mixture of semantics.

We chose three benchmark datasets with part annotations for training and testing, including the ILSVRC 2013 DET Animal-Part dataset~\cite{CNNAoG}, the CUB200-2011 dataset~\cite{CUB200}, and the VOC Part dataset~\cite{SemanticPart}. These datasets provide ground-truth bounding boxes of entire objects. For landmark annotations, the ILSVRC 2013 DET Animal-Part dataset~\cite{CNNAoG} contains ground-truth bounding boxes of heads and legs of 30 animal categories. The CUB200-2011 dataset~\cite{CUB200} contains a total of 11.8K bird images of 200 species, and the dataset provides center positions of 15 bird landmarks. The VOC Part dataset~\cite{SemanticPart} contains ground-truth part segmentations of 107 object landmarks in six animal categories.

We used these datasets, because they contain ground-truth annotations of object landmarks\footnote[2]{To avoid ambiguity, a landmark is referred to as the \textit{central position} of a semantic part (a part with an explicit name, \emph{e.g.} a head, a tail). In contrast, the part corresponding to a filter does not have an explicit name.} (parts) to evaluate the semantic clarity of each filter. As mentioned in \cite{SemanticPart,CNNAoG}, animals usually consist of non-rigid parts, which present considerable challenges for part localization. As in \cite{SemanticPart,CNNAoG}, we selected animal categories in the three datasets for testing.

We learned interpretable filters based on structures of four typical CNNs for evaluation, including the AlexNet~\cite{CNNImageNet}, the VGG-M~\cite{VGG}, the VGG-S~\cite{VGG}, the VGG-16~\cite{VGG}. Note that skip connections in residual networks~\cite{ResNet} make a single feature map contain patterns of different filters. Thus, we did not use residual networks for testing to simplify the evaluation. Given a CNN, all filters in the top conv-layer were set as interpretable filters. Then, we inserted another conv-layer with $M$ filters above the top conv-layer, which did not change the size of output feature maps. \emph{I.e.} we set $M=512$ for the VGG-16, VGG-M, and VGG-S networks, and $M=256$ for the AlexNet. Filters in the new conv-layer were also interpretable filters. Each filter was a $3\times3\times M$ tensor with a bias term.

\begin{figure*}[t]
\centering
\includegraphics[width=\linewidth]{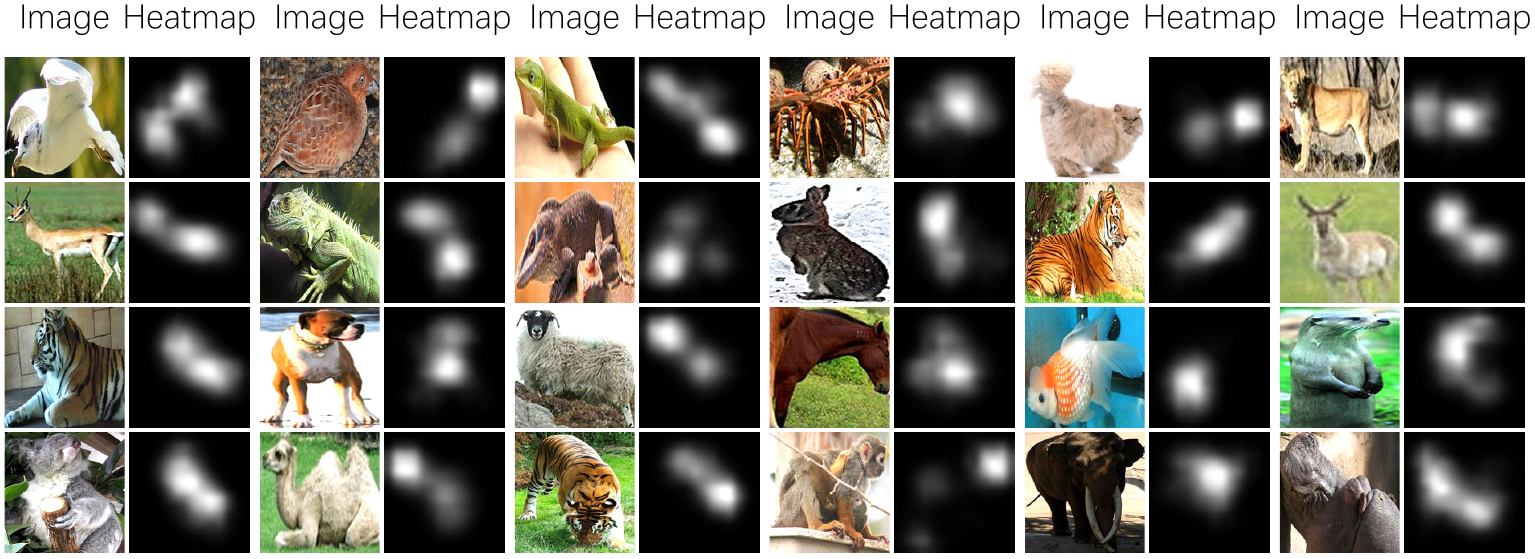}
\caption{Heatmaps for distributions of object parts that are encoded in interpretable filters. We use all filters in the top conv-layer to compute the heatmap. Interpretable filters usually selectively modeled distinct object parts of a category and ignored other parts.}
\label{fig:heatmap}
\end{figure*}

\textit{Implementation details:} We set parameters as $\tau=\frac{0.5}{n^2}$, $\alpha=\frac{n^2}{1+n^2}$, and $\beta\approx4$. $\beta$ was updated during the learning process. We set a decreasing weight for filter losses, \emph{i.e.} $\lambda\propto\frac{1}{t}\mathbb{E}_{x\in{\bf X}}\max_{i,j}x_{ij}$ for the $t$-th epoch. We initialized fully-connected (FC) layers and the new conv-layer, but we loaded parameters of the lower conv-layers from a CNN that was pre-trained using \cite{CNNImageNet,VGG}. We then fine-tuned parameters of all layers in the interpretable CNN using training images in the dataset. To enable a fair comparison, when we learned the traditional CNN as a baseline, we also initialized FC layers of the traditional CNN, used pre-trained parameters in conv-layers, and then fine-tuned the CNN.

\subsection{Experiments}

\textit{Binary classification of a single category:} We learned interpretable CNNs based on above four types of network structures to classify each animal category in above three datasets. We also learned ordinary CNNs using the same data for comparison. We used the logistic log loss for binary classification of a single category from random images. We followed experimental settings in \cite{CNNAoG,explanatoryGraph} to crop objects of the target category as positive samples. Images of other categories were regarded as negative samples.

\textit{Multi-category classification:} We learned interpretable CNNs to classify the six animal categories in the VOC Part dataset~\cite{SemanticPart} and also learned interpretable CNNs to classify the thirty categories in the ILSVRC 2013 DET Animal-Part dataset~\cite{CNNAoG}. In experiments, we tried both the softmax log loss and the logistic log loss\footnote[3]{We considered the output $y_{c}$ for each category $c$ independent to outputs for other categories, thereby a CNN making multiple independent binary classifications of different categories for each image. Table~\ref{tab:classification} reported the average accuracy of the multiple classification outputs of an image.} for multi-category classification.

\subsection{Qualitative Visualization of filters}

We followed the method proposed by Zhou~\emph{et al.}~\cite{CNNSemanticDeep} to compute the receptive fields (RFs) of neural activations of a filter. We used neural activations after ReLU and mask operations and scaled up RFs to the image resolution. As discussed in \cite{Interpretability}, the traditional idea of directly propagating the theoretical receptive field of a neural unit in a feature map back to the image plane cannot accurately reflect the real image-resolution RF of the neural unit (\emph{i.e.} the image region that contributes most to the score of the neural unit). Therefore, we used the method of \cite{CNNSemanticDeep} to compute real RFs.

Studies in both \cite{CNNSemanticDeep} and \cite{Interpretability} have introduced methods to compute real RFs of neural activations on a given feature map. For ordinary CNNs, we simply used a round RF for each neural activation. We overlapped all activated RFs in a feature map to compute the final RF of the feature map.

Fig.~\ref{fig:visual} shows RFs\footnotemark[4] of filters in top conv-layers of CNNs, which were trained for binary classification of a single category. Filters in interpretable CNNs were mainly activated by a certain object part, whereas feature maps of ordinary CNNs after ReLU operations usually represented various object parts and textures. The clear disentanglement of object-part representations can help people to quantify the contribution of different object parts to the network prediction. Fig.~\ref{fig:visual} shows the explanation for part contribution, which was generated by the method of \cite{SemanticalQuantitative}.

We found that interpretable CNNs usually encoded head patterns of animals in its top conv-layer for classification, although no part annotations were used to train the CNN. We can understand such results from the perspective of the information bottleneck~\cite{InformationBottleneck} as follows. (i) Our interpretable filters selectively encode the most distinct parts of each category (\emph{i.e.} the head for most categories), which minimizes the conditional entropy of the final classification given feature maps of a conv-layer. (ii) Each interpretable filter represents a specific part of an object, which minimizes the mutual information between the input image and middle-layer feature maps. The interpretable CNN ``forgets'' as much irrelevant information as possible.

\begin{figure}[t]
\centering
\includegraphics[width=\linewidth]{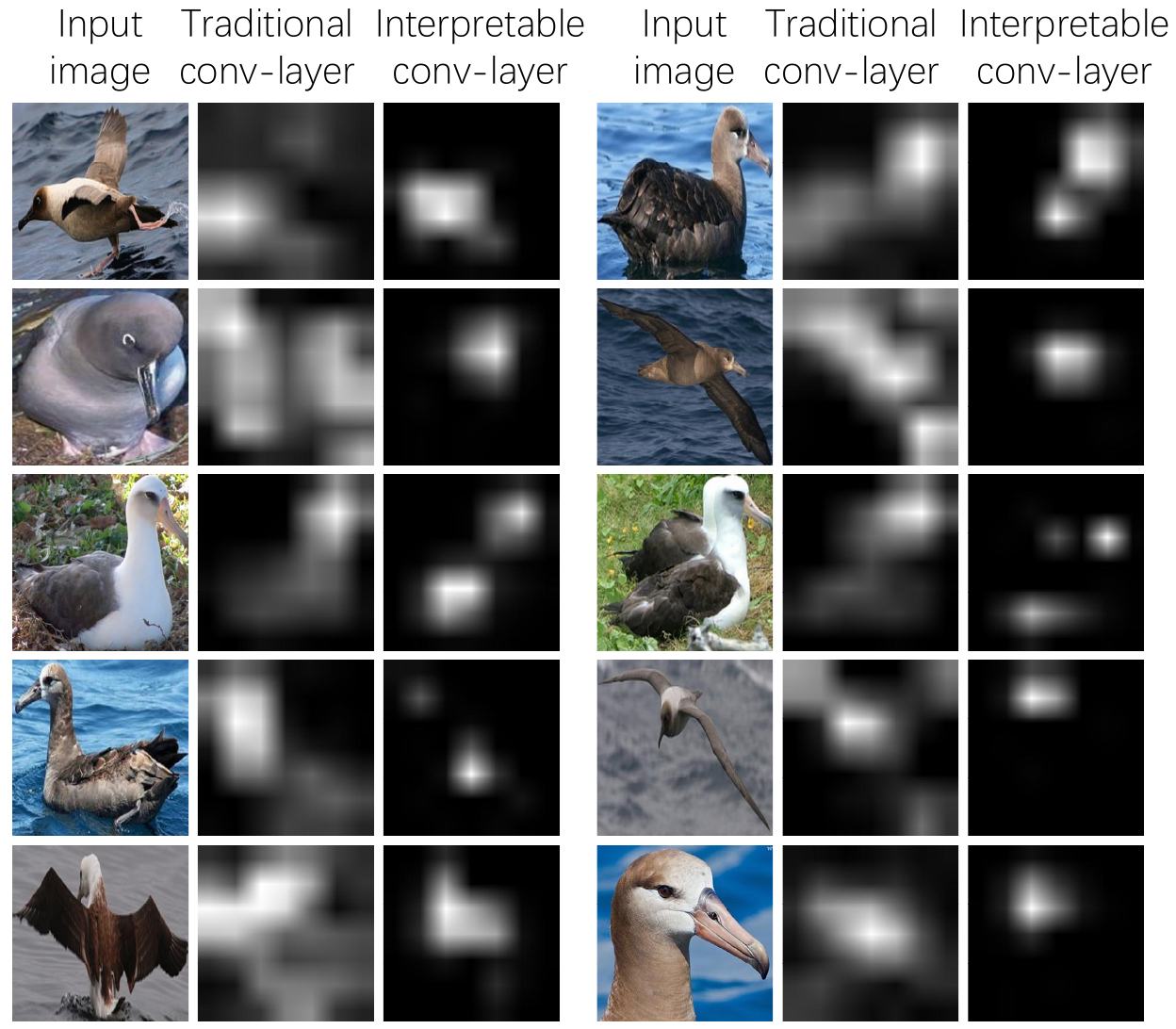}
\caption{Grad-CAM visualizations~\cite{visualCNN_grad_2} of the traditional conv-layer and the interpretable conv-layer. Unlike the traditional conv-layer, the interpretable conv-layer usually selectively modeled distinct object parts of a category and ignored other parts.}
\label{fig:gradCAM}
\end{figure}

In addition to the visualization of RFs, we also visualized heatmaps for part distributions and the grad-CAM attention map of an interpretable conv-layer. Fig.~\ref{fig:heatmap} shows heatmaps for distributions of object parts that were encoded in interpretable filters. Fig.~\ref{fig:gradCAM} compares grad-CAM visualizations~\cite{visualCNN_grad_2} of an interpretable conv-layer and those of a traditional conv-layer. We chose the top conv-layer of the traditional VGG-16 net and the top conv-layer of the interpretable VGG-16 net for visualization. Interpretable filters usually selectively modeled distinct object parts of a category and ignored other parts.

\subsection{Quantitative evaluation of part interpretability}

Filters in low conv-layers usually represent simple patterns or object details, whereas those in high conv-layers are more likely to describe large-scale parts. Therefore, in experiments, we used the following two metrics to evaluate the clarity of part semantics of the top conv-layer of a CNN.

\subsubsection{Evaluation metric: part interpretability}
\label{sec:interpretability}

The metric was originally proposed by Bau~\emph{et al.}~\cite{Interpretability} to measure the object-part interpretability of filters. For each filter $f$, ${\bf X}$ denotes a set of feature maps after ReLU/mask operations on different input images. Then, the distribution of activation scores over all positions in all feature maps was computed. \cite{Interpretability} set a threshold $T_{f}$ such that $p(x_{ij}>T_{f})=0.005$ to select strongest activations from all positions $[i,j]$ from $x\in{\bf X}$ as valid activations for $f$'s semantics.

Then, image-resolution RFs of valid neural activations of each input image $I$ were computed\footnote[4]{\cite{CNNSemanticDeep} computes the RF when the filter represents an object part. Fig.~\ref{fig:visual} used RFs computed by \cite{CNNSemanticDeep} to visualize filters. However, when a filter in an ordinary CNN does not have consistent contours, it is difficult for \cite{CNNSemanticDeep} to align different images to compute an average RF. Thus, for ordinary CNNs, we simply used a round RF for each valid activation. We overlapped all activated RFs in a feature map to compute the final RF as mentioned in \cite{Interpretability}. For a fair comparison, in Section~\ref{sec:interpretability}, we uniformly applied these RFs to both interpretable CNNs and ordinary CNNs.}. The RFs on image $I$, termed $S_{f}^{I}$, corresponded to part regions of $f$.

\begin{table*}
\centering
\resizebox{0.7\linewidth}{!}{\begin{tabular}{c|cccccc|c}
\hline
& bird & cat & cow & dog & horse & sheep & {\bf Avg.}\\
\hline
AlexNet &
0.332&
0.363&
0.340&
0.374&
0.308&
0.373&
0.348
\\
AlexNet, interpretable &
{\bf0.770}&
{\bf0.565}&
{\bf0.618}&
{\bf0.571}&
{\bf0.729}&
{\bf0.669}&
{\bf0.654}
\\
\hline
VGG-16 &
0.519&
0.458&
0.479&
0.534&
0.440&
0.542&
0.495
\\
VGG-16, interpretable &
{\bf0.818}&
{\bf0.653}&
{\bf0.683}&
{\bf0.900}&
{\bf0.795}&
{\bf0.772}&
{\bf0.770}
\\
\hline
VGG-M &
0.357&
0.365&
0.347&
0.368&
0.331&
0.373&
0.357
\\
VGG-M, interpretable &
{\bf0.821}&
{\bf0.632}&
{\bf0.634}&
{\bf0.669}&
{\bf0.736}&
{\bf0.756}&
{\bf0.708}
\\
\hline
VGG-S &
0.251&
0.269&
0.235&
0.275&
0.223&
{\bf0.287}&
0.257
\\
VGG-S, interpretable &
{\bf0.526}&
{\bf0.366}&
{\bf0.291}&
{\bf0.432}&
{\bf0.478}&
0.251&
{\bf0.390}
\\
\hline
\end{tabular}}
\vspace{1pt}
\caption{Part interpretability of filters in CNNs for binary classification of a single category based on the VOC Part dataset~\cite{SemanticPart}.}
\label{tab:interpretability}
\end{table*}

The fitness between the filter $f$ and the $k$-th part on image $I$ was reported as the intersection-over-union score $IoU_{f,k}^{I}=\frac{\Vert S_{f}^{I}\cap S_{k}^{I}\Vert}{\Vert S_{f}^{I}\cup S_{k}^{I}\Vert}$, where $S_{k}^{I}$ represents the ground-truth mask of the $k$-th part on image $I$. Given an image $I$, the filter $f$ was associated with the $k$-th part if $IoU_{f,k}^{I}>0.2$. The criterion $IoU_{f,k}^{I}>0.2$ was stricter than $IoU_{f,k}^{I}>0.04$ in \cite{Interpretability}, because object-part semantics usually needs a stricter criterion than textural semantics and color semantics in \cite{Interpretability}. The average probability of the $k$-th part being associating with the filter $f$ was reported as $P_{f,k}=\mathbb{E}_{I:\textrm{with k-th part}}{\bf1}(IoU_{f,k}^{I}>0.2)$. Note that a single filter may be associated with multiple object parts in an image. The highest probability of part association for each filter was used as the interpretability of filter $f$, \emph{i.e.} $P_{f}=\max_{k}P_{f,k}$.

For the binary classification of a single category, we used testing images of the target category to evaluate the feature interpretability. In the VOC Part dataset~\cite{SemanticPart}, four parts were chosen for the \textit{bird} category. We merged segments of the head, beak, and l/r-eyes as the head part, merged segments of the torso, neck, and l/r-wings as the torso part, merged segments of l/r-legs/feet as the leg part, and used the tail segment as the fourth part. We used five parts for both the \textit{cat} category and the \textit{dog} category. We merged segments of the head, l/r-eyes, l/r-ears, and nose as the head part, merged segments of the torso and neck as the torso part, merged segments of frontal l/r-legs/paws as the frontal legs, merged segments of back l/r-legs/paws as the back legs, and used the tail as the fifth part. Part definitions for the \textit{cow}, \textit{horse}, and \textit{sheep} category were similar those for the cat category, except for that we omitted the tail part of these categories. In particular, we added l/r-horn segments of the horse to the head part. The average part interpretability $P_{f}$ over all filters was computed for evaluation.

For the multi-category classification, we first determined the target category $\hat{c}$ for each filter $f$ \emph{i.e.} $\hat{c}={\arg\!\max}_{c}\mathbb{E}_{x=f(I):I\in{\bf I}_{c}}\sum_{i,j}x_{ij}$. Then, we computed $f$'s object-part interpretability using images of the target category $\hat{c}$ by following above instructions.

\begin{table}[t]
\centering
\resizebox{1.0\linewidth}{!}{\begin{tabular}{ccc}
\hline
Network & Logistic log loss\footnotemark[3] & Softmax log loss\\
\hline
VGG-16 &0.710 &0.723\\
interpretable &{\bf0.938} & {\bf0.897}\\
\hline
VGG-M &0.478 &0.502\\
interpretable &{\bf0.770} &{\bf0.734}\\
\hline
VGG-S &0.479 &0.435\\
interpretable &{\bf0.572} &{\bf0.601}\\
\hline
\end{tabular}}
\vspace{1pt}
\caption{Part interpretability of filters in CNNs that are trained for multi-category classification based on the VOC Part dataset~\cite{SemanticPart}. Filters in our interpretable CNNs exhibited significantly better part interpretability than ordinary CNNs in all comparisons.}
\label{tab:multi-interpretability}
\end{table}

\begin{table*}[t]
\centering
\resizebox{\linewidth}{!}{\begin{tabular}{p{3cm}|ccccccccccc}
\hline
\!\!\!&\!\! gold. \!\!\!&\!\! bird \!\!\!&\!\! frog \!\!\!&\!\! turt. \!\!\!&\!\! liza. \!\!\!&\!\! koala \!\!\!&\!\! lobs. \!\!\!&\!\! dog \!\!\!&\!\! fox \!\!\!&\!\! cat \!\!\!&\!\! lion\\
\!\!\! AlexNet\!\!\!&\!\!
0.161\!\!\!&\!\!
0.167\!\!\!&\!\!
0.152\!\!\!&\!\!
0.153\!\!\!&\!\!
0.175\!\!\!&\!\!
0.128\!\!\!&\!\!
0.123\!\!\!&\!\!
0.144\!\!\!&\!\!
0.143\!\!\!&\!\!
0.148\!\!\!&\!\!
0.137
\\
\!\!\! {\footnotesize AlexNet+ordinary layer}\!\!\!&\!\!
0.154\!\!\!&\!\!
0.157\!\!\!&\!\!
0.143\!\!\!&\!\!
0.146\!\!\!&\!\!
0.170\!\!\!&\!\!
0.120\!\!\!&\!\!
0.118\!\!\!&\!\!
0.127\!\!\!&\!\!
0.117\!\!\!&\!\!
0.136\!\!\!&\!\!
0.120
\\
\!\!\! {\footnotesize AlexNet, interpretable}\!\!\!&\!\!
{\bf0.084}\!\!\!&\!\!
{\bf0.095}\!\!\!&\!\!
{\bf0.090}\!\!\!&\!\!
{\bf0.107}\!\!\!&\!\!
{\bf0.097}\!\!\!&\!\!
{\bf0.079}\!\!\!&\!\!
{\bf0.077}\!\!\!&\!\!
{\bf0.093}\!\!\!&\!\!
{\bf0.087}\!\!\!&\!\!
{\bf0.095}\!\!\!&\!\!
{\bf0.084}
\\
\cline{1-1}
\!\!\! VGG-16\!\!\!&\!\!
0.153\!\!\!&\!\!
0.156\!\!\!&\!\!
0.144\!\!\!&\!\!
0.150\!\!\!&\!\!
0.170\!\!\!&\!\!
0.127\!\!\!&\!\!
0.126\!\!\!&\!\!
0.143\!\!\!&\!\!
0.137\!\!\!&\!\!
0.148\!\!\!&\!\!
0.139
\\
\!\!\! {\footnotesize VGG-16+ordinary layer}\!\!\!&\!\!
0.136\!\!\!&\!\!
0.127\!\!\!&\!\!
0.120\!\!\!&\!\!
0.136\!\!\!&\!\!
0.147\!\!\!&\!\!
0.108\!\!\!&\!\!
0.111\!\!\!&\!\!
0.111\!\!\!&\!\!
0.097\!\!\!&\!\!
0.134\!\!\!&\!\!
0.102
\\
\!\!\! {\footnotesize VGG-16, interpretable}\!\!\!&\!\!
{\bf0.076}\!\!\!&\!\!
{\bf0.099}\!\!\!&\!\!
{\bf0.086}\!\!\!&\!\!
{\bf0.115}\!\!\!&\!\!
{\bf0.113}\!\!\!&\!\!
{\bf0.070}\!\!\!&\!\!
{\bf0.084}\!\!\!&\!\!
{\bf0.077}\!\!\!&\!\!
{\bf0.069}\!\!\!&\!\!
{\bf0.086}\!\!\!&\!\!
{\bf0.067}
\\
\cline{1-1}
\!\!\! VGG-M\!\!\!&\!\!
0.161\!\!\!&\!\!
0.166\!\!\!&\!\!
0.151\!\!\!&\!\!
0.153\!\!\!&\!\!
0.176\!\!\!&\!\!
0.128\!\!\!&\!\!
0.125\!\!\!&\!\!
0.145\!\!\!&\!\!
0.145\!\!\!&\!\!
0.150\!\!\!&\!\!
0.140
\\
\!\!\! {\footnotesize VGG-M+ordinary layer}\!\!\!&\!\!
0.147\!\!\!&\!\!
0.144\!\!\!&\!\!
0.135\!\!\!&\!\!
0.142\!\!\!&\!\!
0.159\!\!\!&\!\!
0.114\!\!\!&\!\!
0.115\!\!\!&\!\!
0.119\!\!\!&\!\!
0.111\!\!\!&\!\!
0.128\!\!\!&\!\!
0.114
\\
\!\!\! {\footnotesize VGG-M, interpretable}\!\!\!&\!\!
{\bf0.088}\!\!\!&\!\!
{\bf0.088}\!\!\!&\!\!
{\bf0.089}\!\!\!&\!\!
{\bf0.108}\!\!\!&\!\!
{\bf0.099}\!\!\!&\!\!
{\bf0.080}\!\!\!&\!\!
{\bf0.074}\!\!\!&\!\!
{\bf0.090}\!\!\!&\!\!
{\bf0.082}\!\!\!&\!\!
{\bf0.103}\!\!\!&\!\!
{\bf0.079}
\\
\cline{1-1}
\!\!\! VGG-S\!\!\!&\!\!
0.158\!\!\!&\!\!
0.166\!\!\!&\!\!
0.149\!\!\!&\!\!
0.151\!\!\!&\!\!
0.173\!\!\!&\!\!
0.127\!\!\!&\!\!
0.124\!\!\!&\!\!
0.143\!\!\!&\!\!
0.142\!\!\!&\!\!
0.148\!\!\!&\!\!
0.138
\\
\!\!\! {\footnotesize VGG-S+ordinary layer}\!\!\!&\!\!
0.150\!\!\!&\!\!
0.132\!\!\!&\!\!
0.133\!\!\!&\!\!
0.138\!\!\!&\!\!
0.156\!\!\!&\!\!
0.113\!\!\!&\!\!
0.111\!\!\!&\!\!
0.110\!\!\!&\!\!
0.104\!\!\!&\!\!
0.125\!\!\!&\!\!
0.112
\\
\!\!\! {\footnotesize VGG-S, interpretable}\!\!\!&\!\!
{\bf0.087}\!\!\!&\!\!
{\bf0.101}\!\!\!&\!\!
{\bf0.093}\!\!\!&\!\!
{\bf0.107}\!\!\!&\!\!
{\bf0.096}\!\!\!&\!\!
{\bf0.084}\!\!\!&\!\!
{\bf0.078}\!\!\!&\!\!
{\bf0.091}\!\!\!&\!\!
{\bf0.082}\!\!\!&\!\!
{\bf0.101}\!\!\!&\!\!
{\bf0.082}
\\
\hline
\!\!\!&\!\! tiger \!\!\!&\!\! bear \!\!\!&\!\! rabb. \!\!\!&\!\! hams. \!\!\!&\!\! squi. \!\!\!&\!\! horse \!\!\!&\!\! zebra \!\!\!&\!\! swine \!\!\!&\!\! hippo. \!\!\!&\!\! catt. \!\!\!&\!\! sheep\\
\!\!\! AlexNet\!\!\!&\!\!
0.142\!\!\!&\!\!
0.144\!\!\!&\!\!
0.148\!\!\!&\!\!
0.128\!\!\!&\!\!
0.149\!\!\!&\!\!
0.152\!\!\!&\!\!
0.154\!\!\!&\!\!
0.141\!\!\!&\!\!
0.141\!\!\!&\!\!
0.144\!\!\!&\!\!
0.155
\\
\!\!\! {\footnotesize AlexNet+ordinary layer}\!\!\!&\!\!
0.123\!\!\!&\!\!
0.133\!\!\!&\!\!
0.136\!\!\!&\!\!
0.112\!\!\!&\!\!
0.145\!\!\!&\!\!
0.149\!\!\!&\!\!
0.142\!\!\!&\!\!
0.137\!\!\!&\!\!
0.139\!\!\!&\!\!
0.141\!\!\!&\!\!
0.149
\\
\!\!\! {\footnotesize AlexNet, interpretable}\!\!\!&\!\!
{\bf0.090}\!\!\!&\!\!
{\bf0.095}\!\!\!&\!\!
{\bf0.095}\!\!\!&\!\!
{\bf0.077}\!\!\!&\!\!
{\bf0.095}\!\!\!&\!\!
{\bf0.098}\!\!\!&\!\!
{\bf0.084}\!\!\!&\!\!
{\bf0.091}\!\!\!&\!\!
{\bf0.089}\!\!\!&\!\!
{\bf0.097}\!\!\!&\!\!
{\bf0.101}
\\
\cline{1-1}
\!\!\! VGG-16\!\!\!&\!\!
0.144\!\!\!&\!\!
0.143\!\!\!&\!\!
0.146\!\!\!&\!\!
0.125\!\!\!&\!\!
0.150\!\!\!&\!\!
0.150\!\!\!&\!\!
0.153\!\!\!&\!\!
0.141\!\!\!&\!\!
0.140\!\!\!&\!\!
0.140\!\!\!&\!\!
0.150
\\
\!\!\! {\footnotesize VGG-16+ordinary layer}\!\!\!&\!\!
0.127\!\!\!&\!\!
0.112\!\!\!&\!\!
0.119\!\!\!&\!\!
0.100\!\!\!&\!\!
0.112\!\!\!&\!\!
0.134\!\!\!&\!\!
0.140\!\!\!&\!\!
0.126\!\!\!&\!\!
0.126\!\!\!&\!\!
0.131\!\!\!&\!\!
0.135
\\
\!\!\! {\footnotesize VGG-16, interpretable}\!\!\!&\!\!
{\bf0.097}\!\!\!&\!\!
{\bf0.081}\!\!\!&\!\!
{\bf0.079}\!\!\!&\!\!
{\bf0.066}\!\!\!&\!\!
{\bf0.065}\!\!\!&\!\!
{\bf0.106}\!\!\!&\!\!
{\bf0.077}\!\!\!&\!\!
{\bf0.094}\!\!\!&\!\!
{\bf0.083}\!\!\!&\!\!
{\bf0.102}\!\!\!&\!\!
{\bf0.097}
\\
\cline{1-1}
\!\!\! VGG-M\!\!\!&\!\!
0.145\!\!\!&\!\!
0.144\!\!\!&\!\!
0.150\!\!\!&\!\!
0.128\!\!\!&\!\!
0.150\!\!\!&\!\!
0.151\!\!\!&\!\!
0.158\!\!\!&\!\!
0.140\!\!\!&\!\!
0.140\!\!\!&\!\!
0.143\!\!\!&\!\!
0.155
\\
\!\!\! {\footnotesize VGG-M+ordinary layer}\!\!\!&\!\!
0.124\!\!\!&\!\!
0.131\!\!\!&\!\!
0.134\!\!\!&\!\!
0.108\!\!\!&\!\!
0.132\!\!\!&\!\!
0.138\!\!\!&\!\!
0.141\!\!\!&\!\!
0.133\!\!\!&\!\!
0.131\!\!\!&\!\!
0.135\!\!\!&\!\!
0.142
\\
\!\!\! {\footnotesize VGG-M, interpretable}\!\!\!&\!\!
{\bf0.089}\!\!\!&\!\!
{\bf0.101}\!\!\!&\!\!
{\bf0.097}\!\!\!&\!\!
{\bf0.082}\!\!\!&\!\!
{\bf0.095}\!\!\!&\!\!
{\bf0.095}\!\!\!&\!\!
{\bf0.080}\!\!\!&\!\!
{\bf0.095}\!\!\!&\!\!
{\bf0.084}\!\!\!&\!\!
{\bf0.092}\!\!\!&\!\!
{\bf0.094}
\\
\cline{1-1}
\!\!\! VGG-S\!\!\!&\!\!
0.142\!\!\!&\!\!
0.143\!\!\!&\!\!
0.148\!\!\!&\!\!
0.128\!\!\!&\!\!
0.146\!\!\!&\!\!
0.149\!\!\!&\!\!
0.155\!\!\!&\!\!
0.139\!\!\!&\!\!
0.140\!\!\!&\!\!
0.141\!\!\!&\!\!
0.155
\\
\!\!\! {\footnotesize VGG-S+ordinary layer}\!\!\!&\!\!
0.117\!\!\!&\!\!
0.127\!\!\!&\!\!
0.127\!\!\!&\!\!
0.105\!\!\!&\!\!
0.122\!\!\!&\!\!
0.136\!\!\!&\!\!
0.137\!\!\!&\!\!
0.133\!\!\!&\!\!
0.131\!\!\!&\!\!
0.130\!\!\!&\!\!
0.143
\\
\!\!\! {\footnotesize VGG-S, interpretable}\!\!\!&\!\!
{\bf0.089}\!\!\!&\!\!
{\bf0.097}\!\!\!&\!\!
{\bf0.091}\!\!\!&\!\!
{\bf0.076}\!\!\!&\!\!
{\bf0.098}\!\!\!&\!\!
{\bf0.096}\!\!\!&\!\!
{\bf0.080}\!\!\!&\!\!
{\bf0.092}\!\!\!&\!\!
{\bf0.088}\!\!\!&\!\!
{\bf0.094}\!\!\!&\!\!
{\bf0.101}
\\
\hline
\!\!\!&\!\! ante. \!\!\!&\!\! camel \!\!\!&\!\! otter \!\!\!&\!\! arma. \!\!\!&\!\! monk. \!\!\!&\!\! elep. \!\!\!&\!\! red pa. \!\!\!&\!\! gia.pa. \!\!\!&\!\! \!\!\!&\!\! \!\!\!&\!\! \textcolor{blue}{\bf Avg.}\\
\!\!\! AlexNet\!\!\!&\!\!
0.147\!\!\!&\!\!
0.153\!\!\!&\!\!
0.159\!\!\!&\!\!
0.160\!\!\!&\!\!
0.139\!\!\!&\!\!
0.125\!\!\!&\!\!
0.140\!\!\!&\!\!
0.125\!\!\!&\!\!
\!\!\!&\!\!
\!\!\!&\!\!
\textcolor{blue}{0.146}
\\
\!\!\! {\footnotesize AlexNet+ordinary layer}\!\!\!&\!\!
0.148\!\!\!&\!\!
0.143\!\!\!&\!\!
0.145\!\!\!&\!\!
0.151\!\!\!&\!\!
0.125\!\!\!&\!\!
0.116\!\!\!&\!\!
0.127\!\!\!&\!\!
0.102\!\!\!&\!\!
\!\!\!&\!\!
\!\!\!&\!\!
\textcolor{blue}{0.136}
\\
\!\!\! {\footnotesize AlexNet, interpretable}\!\!\!&\!\!
{\bf0.085}\!\!\!&\!\!
{\bf0.102}\!\!\!&\!\!
{\bf0.104}\!\!\!&\!\!
{\bf0.095}\!\!\!&\!\!
{\bf0.090}\!\!\!&\!\!
{\bf0.085}\!\!\!&\!\!
{\bf0.084}\!\!\!&\!\!
{\bf0.073}\!\!\!&\!\!
\!\!\!&\!\!
\!\!\!&\!\!
\textcolor{blue}{\bf0.091}
\\
\cline{1-1}
\!\!\! VGG-16\!\!\!&\!\!
0.144\!\!\!&\!\!
0.149\!\!\!&\!\!
0.154\!\!\!&\!\!
0.163\!\!\!&\!\!
0.136\!\!\!&\!\!
0.129\!\!\!&\!\!
0.143\!\!\!&\!\!
0.125\!\!\!&\!\!
\!\!\!&\!\!
\!\!\!&\!\!
\textcolor{blue}{0.144}
\\
\!\!\! {\footnotesize VGG-16+ordinary layer}\!\!\!&\!\!
0.122\!\!\!&\!\!
0.121\!\!\!&\!\!
0.134\!\!\!&\!\!
0.143\!\!\!&\!\!
0.108\!\!\!&\!\!
0.110\!\!\!&\!\!
0.115\!\!\!&\!\!
0.102\!\!\!&\!\!
\!\!\!&\!\!
\!\!\!&\!\!
\textcolor{blue}{0.121}
\\
\!\!\! {\footnotesize VGG-16, interpretable}\!\!\!&\!\!
{\bf0.091}\!\!\!&\!\!
{\bf0.105}\!\!\!&\!\!
{\bf0.093}\!\!\!&\!\!
{\bf0.100}\!\!\!&\!\!
{\bf0.074}\!\!\!&\!\!
{\bf0.084}\!\!\!&\!\!
{\bf0.067}\!\!\!&\!\!
{\bf0.063}\!\!\!&\!\!
\!\!\!&\!\!
\!\!\!&\!\!
\textcolor{blue}{\bf0.085}
\\
\cline{1-1}
\!\!\! VGG-M\!\!\!&\!\!
0.146\!\!\!&\!\!
0.154\!\!\!&\!\!
0.160\!\!\!&\!\!
0.161\!\!\!&\!\!
0.140\!\!\!&\!\!
0.126\!\!\!&\!\!
0.142\!\!\!&\!\!
0.127\!\!\!&\!\!
\!\!\!&\!\!
\!\!\!&\!\!
\textcolor{blue}{0.147}
\\
\!\!\! {\footnotesize VGG-M+ordinary layer}\!\!\!&\!\!
0.130\!\!\!&\!\!
0.135\!\!\!&\!\!
0.140\!\!\!&\!\!
0.150\!\!\!&\!\!
0.120\!\!\!&\!\!
0.112\!\!\!&\!\!
0.120\!\!\!&\!\!
0.106\!\!\!&\!\!
\!\!\!&\!\!
\!\!\!&\!\!
\textcolor{blue}{0.130}
\\
\!\!\! {\footnotesize VGG-M, interpretable}\!\!\!&\!\!
{\bf0.077}\!\!\!&\!\!
{\bf0.104}\!\!\!&\!\!
{\bf0.102}\!\!\!&\!\!
{\bf0.093}\!\!\!&\!\!
{\bf0.086}\!\!\!&\!\!
{\bf0.087}\!\!\!&\!\!
{\bf0.089}\!\!\!&\!\!
{\bf0.068}\!\!\!&\!\!
\!\!\!&\!\!
\!\!\!&\!\!
\textcolor{blue}{\bf0.090}
\\
\cline{1-1}
\!\!\! VGG-S\!\!\!&\!\!
0.143\!\!\!&\!\!
0.154\!\!\!&\!\!
0.158\!\!\!&\!\!
0.157\!\!\!&\!\!
0.140\!\!\!&\!\!
0.125\!\!\!&\!\!
0.139\!\!\!&\!\!
0.125\!\!\!&\!\!
\!\!\!&\!\!
\!\!\!&\!\!
\textcolor{blue}{0.145}
\\
\!\!\! {\footnotesize VGG-S+ordinary layer}\!\!\!&\!\!
0.125\!\!\!&\!\!
0.133\!\!\!&\!\!
0.135\!\!\!&\!\!
0.147\!\!\!&\!\!
0.119\!\!\!&\!\!
0.111\!\!\!&\!\!
0.118\!\!\!&\!\!
0.100\!\!\!&\!\!
\!\!\!&\!\!
\!\!\!&\!\!
\textcolor{blue}{0.126}
\\
\!\!\! {\footnotesize VGG-S, interpretable}\!\!\!&\!\!
{\bf0.077}\!\!\!&\!\!
{\bf0.102}\!\!\!&\!\!
{\bf0.105}\!\!\!&\!\!
{\bf0.094}\!\!\!&\!\!
{\bf0.090}\!\!\!&\!\!
{\bf0.086}\!\!\!&\!\!
{\bf0.078}\!\!\!&\!\!
{\bf0.072}\!\!\!&\!\!
\!\!\!&\!\!
\!\!\!&\!\!
\textcolor{blue}{\bf0.090}
\\
\hline
\end{tabular}}
\vspace{1pt}
\caption{Location instability of filters ($\mathbb{E}_{f,k}[D_{f,k}]$) in CNNs that are trained for the binary classification of a single category using the ILSVRC 2013 DET Animal-Part dataset~\cite{CNNAoG}. Filters in our interpretable CNNs exhibited significantly lower localization instability than ordinary CNNs.}
\label{tab:imgnet-stability}
\end{table*}

\subsubsection{Evaluation metric: location instability}

The second metric measures the instability of part locations, which was used in \cite{explanatoryGraph,interpretableCNN}. It is assumed that if $f$ consistently represented the same object part through different objects, then distances between the inferred part $\hat{\mu}$ and some ground-truth landmarks\footnotemark[2] should keep stable among different objects. For example, if $f$ represented the shoulder part without ambiguity, then the distance between the inferred position and the head will not change a lot among different objects.

Therefore, the deviation of the distance between the inferred position $\hat{\mu}$ and a specific ground-truth landmark among different images was computed. The location $\hat{\mu}$ was inferred as the neural unit with the highest activation on $f$'s feature map. We reported the average deviation \emph{w.r.t.} different landmarks as the location instability of $f$.

\begin{figure}[t]
\centering
\includegraphics[width=\linewidth]{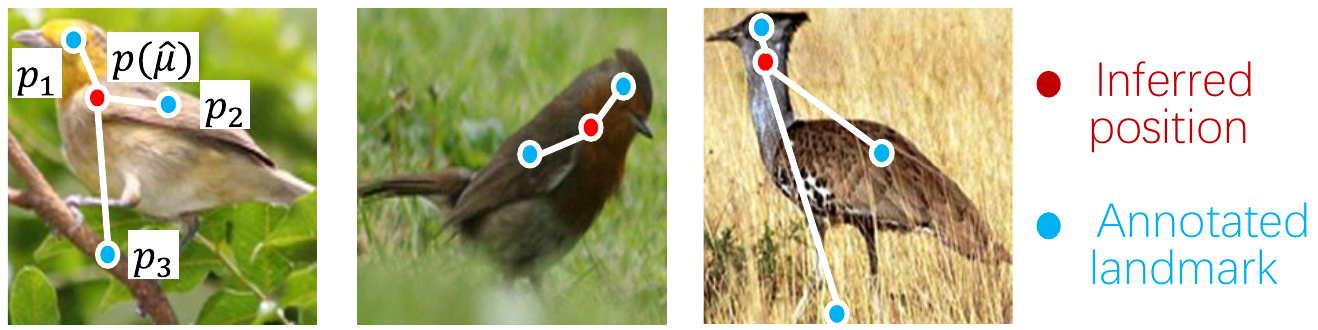}
\caption{Notation for computing the location instability.}
\label{fig:instability}
\end{figure}

Please see Fig.~\ref{fig:instability}. Given an input image $I$, $d_{I}(p_{k},\hat{\mu})=\frac{\Vert {\bf p}_{k}-{\bf p}(\hat{\mu})\Vert}{\sqrt{w^2+h^2}}$ denotes the normalized distance between the inferred part and the $k$-th landmark ${\bf p}_{k}$, where ${\bf p}(\hat{\mu})$ is referred to as the center of the unit $\hat{\mu}$'s RF. $\sqrt{w^2+h^2}$ measures the diagonal length of $I$. $D_{f,k}=\sqrt{{\textrm{var}}_{I}[d_{I}(p_{k},\hat{\mu})]}$ is termed as the \textit{relative location deviation} of filter $f$ \emph{w.r.t.} the $k$-th landmark, where ${\textrm{var}}_{I}[d_{I}(p_{k},\hat{\mu})]$ is the variation of $d_{I}(p_{k},\hat{\mu})$. Because each landmark could not appear in all testing images, for each filter $f$, the metric only used inference results on top-ranked 100 images with the highest inference scores to compute $D_{f,k}$. In this way, the average of relative location deviations of all the filters in a conv-layer \emph{w.r.t.} all $K$ landmarks, \emph{i.e.} $\mathbb{E}_{f}\mathbb{E}_{k=1}^{K}D_{f,k}$, was reported as the location instability of $f$.

We used the most frequent object parts as landmarks to measure the location instability. For the ILSVRC 2013 DET Animal-Part dataset~\cite{CNNAoG}, we used the \textit{head} and \textit{frontal legs} of each category as landmarks for evaluation. For the VOC Part dataset~\cite{SemanticPart}, we selected the \textit{head}, \textit{neck}, and \textit{torso} of each category as landmarks. For the CUB200-2011 dataset~\cite{CUB200}, we used the \textit{head}, \textit{back}, \textit{tail} of birds as landmarks.

\begin{table*}[t]
\centering
\resizebox{0.7\linewidth}{!}{\begin{tabular}{l|cccccc|c}
\hline
& bird & cat & cow & dog & horse & sheep & \textcolor{blue}{\bf Avg.}\\
\hline
AlexNet &
0.153&
0.131&
0.141&
0.128&
0.145&
0.140&
\textcolor{blue}{0.140}
\\
AlexNet+ordinary layer &
0.147&
0.125&
0.139&
0.112&
0.146&
0.143&
\textcolor{blue}{0.136}
\\
AlexNet, interpretable w/o filter loss &0.091&0.090&0.091&0.089&0.086&0.088&\textcolor{blue}{0.089}\\
AlexNet, interpretable &
{0.090}&
{0.089}&
{0.090}&
{0.088}&
{0.087}&
{0.088}&
\textcolor{blue}{\bf0.088}
\\
\hline
VGG-16 &
0.145&
0.133&
0.146&
0.127&
0.143&
0.143&
\textcolor{blue}{0.139}
\\
VGG-16+ordinary layer &
0.125&
0.121&
0.137&
0.102&
0.131&
0.137&
\textcolor{blue}{0.125}
\\
VGG-16, interpretable w/o filter loss &0.099&0.087&0.102&0.078&0.096&0.101&\textcolor{blue}{\bf0.094}\\
VGG-16, interpretable &
{0.101}&
{0.098}&
{0.105}&
{0.074}&
{0.097}&
{0.100}&
\textcolor{blue}{0.096}
\\
\hline
VGG-M &
0.152&
0.132&
0.143&
0.130&
0.145&
0.141&
\textcolor{blue}{0.141}
\\
VGG-M+ordinary layer &
0.142&
0.120&
0.139&
0.115&
0.141&
0.142&
\textcolor{blue}{0.133}
\\
VGG-M, interpretable w/o filter loss &0.089&0.095&0.091&0.086&0.086&0.091&\textcolor{blue}{0.090}\\
VGG-M, interpretable &
{0.086}&
{0.094}&
{0.090}&
{0.087}&
{0.084}&
{0.084}&
\textcolor{blue}{\bf0.088}
\\
\hline
VGG-S &
0.152&
0.131&
0.141&
0.128&
0.144&
0.141&
\textcolor{blue}{0.139}
\\
VGG-S+ordinary layer &
0.137&
0.115&
0.133&
0.107&
0.133&
0.138&
\textcolor{blue}{0.127}
\\
VGG-S, interpretable w/o filter loss &0.093&0.096&0.093&0.086&0.083&0.090&\textcolor{blue}{0.090}\\
VGG-S, interpretable &
{0.089}&
{0.092}&
{0.092}&
{0.087}&
{0.086}&
{0.088}&
\textcolor{blue}{\bf0.089}
\\
\hline
\end{tabular}}
\vspace{1pt}
\caption{Location instability of filters ($\mathbb{E}_{f,k}[D_{f,k}]$) in CNNs that are trained for binary classification of a single category using the VOC Part dataset~\cite{SemanticPart}. Filters in our interpretable CNNs exhibited significantly lower localization instability than ordinary CNNs.}
\label{tab:voc-stability}
\end{table*}

\begin{table}[t]
\centering
\resizebox{\linewidth}{!}{\begin{tabular}{lc}
\hline
\multicolumn{1}{c}{Neural network} & \qquad Avg. location instability\\
\hline
AlexNet &0.150\\
AlexNet+ordinary layer & 0.118\\
AlexNet, interpretable &{\bf0.070}\\
\hline
VGG-16 &0.137\\
VGG-16+ordinary layer & 0.097\\
VGG-16, interpretable &{\bf0.076}\\
\hline
VGG-M &0.148\\
VGG-M+ordinary layer & 0.107\\
VGG-M, interpretable &{\bf0.065}\\
\hline
VGG-S &0.148\\
VGG-S+ordinary layer & 0.103\\
VGG-S, interpretable &{\bf0.073}\\
\hline
\end{tabular}}
\vspace{1pt}
\caption{Location instability of filters ($\mathbb{E}_{f,k}[D_{f,k}]$) in CNNs for binary classification of a single category using the CUB200-2011 dataset.}
\label{tab:cub200-stability}
\end{table}

\begin{table}[t]
\centering
\resizebox{1.0\linewidth}{!}{\begin{tabular}{c|c|cc}
\hline
& {\footnotesize ILSVRC Part~\cite{CNNAoG}} & \multicolumn{2}{|c}{VOC Part~\cite{SemanticPart}}\\
\hline
& {\footnotesize Logistic} & {\footnotesize Logistic} & {\footnotesize Softmax}\\
& {\footnotesize log loss\footnotemark[3]} & {\footnotesize log loss\footnotemark[3]} & {\footnotesize log loss}\\
\hline
VGG-16 & -- &0.128 &0.142\\
{\small ordinary layer} & -- &0.096 &0.099\\
{\small interpretable} & -- &{\bf0.073} &{\bf0.075}\\
\hline
VGG-M & 0.167 &0.135 &0.137\\
{\small ordinary layer} & -- &0.117 &0.107\\
{\small interpretable} &{\bf0.096} &{\bf0.083} &{\bf 0.087}\\
\hline
VGG-S & 0.131 &0.138 &0.138\\
{\small ordinary layer} & -- &0.127 &0.099\\
{\small interpretable} &{\bf0.083} &{\bf0.078} &{\bf0.082}\\
\hline
\end{tabular}}
\vspace{1pt}
\caption{Location instability of filters ($\mathbb{E}_{f,k}[D_{f,k}]$) in CNNs that are trained for multi-category classification. Filters in our interpretable CNNs exhibited significantly lower localization instability than ordinary CNNs in all comparisons.}
\label{tab:multi-stability}
\end{table}

In particular, for multi-category classification, we first determined the target category of for each filter $f$ and then computed the relative location deviation $D_{f,k}$ using landmarks of $f$'s target category. Because filters in baseline CNNs did not exclusively represent a single category, we simply assigned filter $f$ with the category whose landmarks can achieve the lowest location deviation to simplify the computation. \emph{I.e.} for a baseline CNN, we used $\mathbb{E}_{f}\min_{c}\mathbb{E}_{k\in Part_{c}}D_{f,k}$ to evaluate the location instability, where $Part_{c}$ denotes the set of part indexes belonging to category $c$.

\subsubsection{Comparisons between metrics of filter interpretability and location instability}

Although the filter interpretability~\cite{Interpretability} and the location instability~\cite{interpretableCNN} are the two most state-of-the-art metrics to evaluate the interpretability of a convolution filter, these metrics still have some limitations.

Firstly, the filter interpretability~\cite{Interpretability} assumes that the feature map of an automatically learned filter should well match the ground-truth segment of a semantic part (with an explicit part name), an object, or a texture. For example, it assumes that a filter may represent the exact segment of the head part. However, without ground-truth annotations of object parts or textures for supervision, there is no mechanism to assign explicit semantic meanings with filters during the learning process. In most cases, filters in an interpretable CNN (as well as a few filters in traditional CNNs) may describe a specific object part without explicit names, \emph{e.g.} the region of both the head and neck or the region connecting the torso and the tail. Therefore, in both \cite{Interpretability} and \cite{interpretableCNN}, people did not require the inferred object region to describe the exact segment of a semantic part, and simply set a relatively loose criterion $IoU_{f,k}^{I}>0.04\;\textrm{or}\;0.2$ to compute the filter interpretability.

Secondly, the location instability was proposed in \cite{interpretableCNN}. The location instability of a filter is evaluated using the average deviation of distances between the inferred position and some ground-truth landmarks. There is also an assumption for this evaluation metric, \emph{i.e.} the distance between an inferred part and a specific landmark should not change a lot through different images. As a result, people cannot set landmarks as the head and the tail of a snake, because the distance between different parts of a snake continuously changes when the snake moves.

Generally speaking, there are two advantages to use the location instability for evaluation:
\begin{itemize}
\item The computation of the location instability~\cite{interpretableCNN} is independent to the size of the receptive field (RF) of a neural activation. This solves a big problem with the evaluation of filter interpretability, \emph{i.e.} state-of-the-art methods of computing a neural activation's image-resolution RFs (\emph{e.g.} \cite{CNNSemanticDeep}) can only provide an approximate scale of the RF. The metric of location instability only uses central positions of part inferences of a filter, rather than use the entire inferred part segment, for evaluation. Thus, the location instability is a robust metric to evaluate the object-part interpretability of a filter.
\item The location instability allows a filter to represent an object part without an explicit name (a half of the head).
\end{itemize}
Nevertheless, the evaluation metric for filter interpretability is still an open problem.

\subsubsection{Robustness to adversarial attacks}

In this experiment, we applied adversarial attacks~\cite{CNNAnalysis_1} to both original CNNs and interpretable CNNs. The CNNs were learned to classify birds in the CUB200-2011 dataset and random images. Table~\ref{tab:adversarial} compares the average adversarial distortion $\sqrt{\frac{\sum\left(I_{i}^{\prime}-I_{i}\right)^{2}}{n}}$ of the adversarial signal among all images between original CNNs and interpretable CNNs, where $I$ represents the input image while $I'$ denotes the adversarial counterpart. Because interpretable CNNs exclusively encoded object-part patterns and ignored textures, original CNNs usually exhibited stronger robustness to adversarial attacks than interpretable CNNs.

\subsubsection{Experimental results and analysis}

Feature interpretability of different CNNs is evaluated in Tables~\ref{tab:interpretability}, \ref{tab:multi-interpretability}, \ref{tab:imgnet-stability}, \ref{tab:voc-stability}, \ref{tab:cub200-stability}, and \ref{tab:multi-stability}. Tables~\ref{tab:interpretability} and \ref{tab:multi-interpretability} show results based on the metric in \cite{Interpretability}. Tables~\ref{tab:imgnet-stability}, \ref{tab:voc-stability}, and \ref{tab:cub200-stability} list location instability of CNNs for binary classification of a single category. Table~\ref{tab:multi-stability} reports location instability of CNNs that were learned for multi-category classification.

We compared our interpretable CNNs with two types of CNNs, \emph{i.e.} the original CNN, the CNN with an additional conv-layer on the top (termed \textit{AlexNet/VGG-16/VGG-M/VGG-S+ordinary layer}). To construct the CNN with a new conv-layer, we put a new conv-layer on the top of conv-layer. The filter size of the new conv-layer was $3\times3\times channel\,\,number$, and output feature maps of the new conv-layer were in the same size of input feature maps. Because our interpretable CNN had an additional interpretable conv-layer, we designed the baseline CNN with a new conv-layer to enable fair comparisons. Our interpretable filters exhibited significantly higher part interpretability and lower location instability than traditional filters in baseline CNNs over almost all comparisons. Table~\ref{tab:classification} reports the classification accuracy of different CNNs. Ordinary CNNs exhibited better performance in binary classification, while interpretable CNNs outperformed baseline CNNs in multi-category classification.

In addition, to prove the discrimination power of the learned filter, we further tested the average accuracy when we used the maximum activation score in a single filter's feature map as a metric for binary classification between birds in the CUB200-2011 dataset~\cite{CUB200} and random images. In the scenario of classifying birds from random images, filters in the CNN was expected to learn the common appearance of birds, instead of summarizing knowledge from random images. Thus, we chose filters in the top conv-layer. If the maximum activation score of a filter exceeded a threshold, then we classified the input image as a bird; otherwise not. The threshold was set to the one that maximized the classification accuracy. Table~\ref{tab:filterClassify} reports the average classification accuracy over all filters. Our interpretable filters outperformed ordinary filters.

%\textcolor{red}{We further tested effects of the interpretable loss in the neural activations among different categories. We used the VGG-M, VGG-S, and VGG-16 networks with either the logistic log loss or the softmax loss, which was trained to classify animal categories in the VOC Part dataset~\cite{SemanticPart}. For each interpretable filter, given images of its target category, we recorded their neural activations (\emph{i.e.} recording the maximal activation value in each of their feature maps). At the same time, we also recorded neural activations on other categories of the filter. The interpretable filter was supposed to activate much more strongly on its target category than on other (unrelated) categories. We collected all activation records on corresponding categories of all filters, and their mean value is reported in Table~\ref{tab:activation}. In comparison, we also computed the mean value of all activations on unrelated categories of all filters in Table~\ref{tab:activation}. This table shows that the interpretable filter was usually activated more strongly on the target category than on other categories. Furthermore, we also compared the proposed interpretable CNN with the ablation baseline \textit{interpretable w/o filter loss}, in which the CNN was learned without the filter loss. Table~\ref{tab:activation} shows that the filter loss made each filter more prone to being triggered by a single category, \emph{i.e.} boosting the feature interpretability.}

Given a CNN for binary classification of an animal category in the VOC Part dataset~\cite{SemanticPart}, we manually annotated the part name corresponding to the learned filters in the CNN. Table~\ref{tab:semantics} reports the ratio of interpretable filters that corresponds to each object part.

\begin{figure}[t]
\centering
\includegraphics[width=\linewidth]{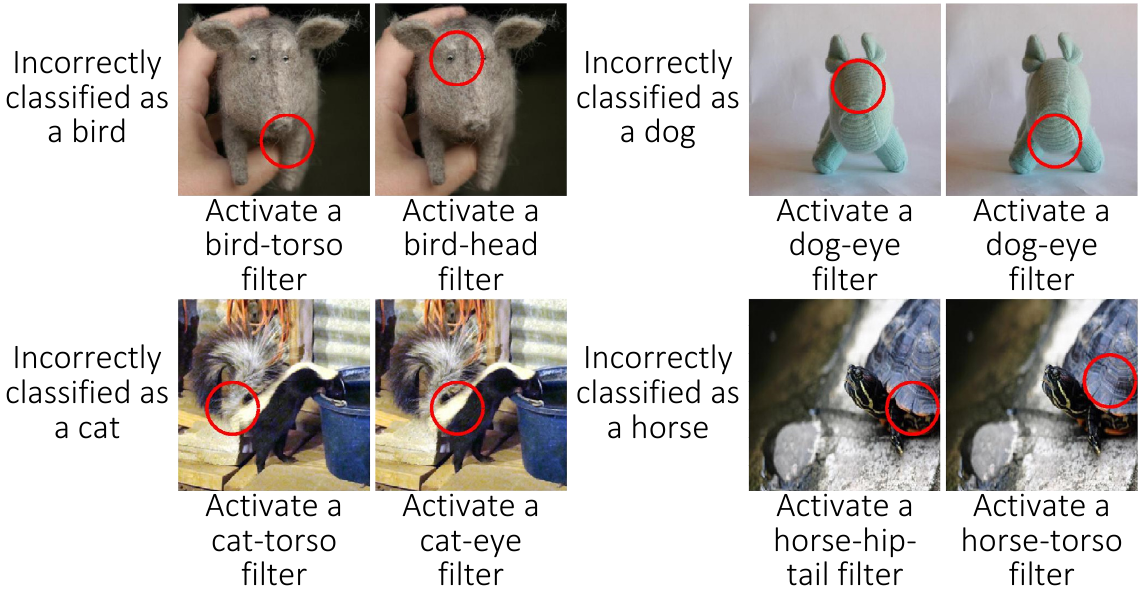}
\caption{Examples that were incorrectly classified by the interpretable CNN.}
\label{fig:miscls}
\end{figure}

Besides, we also analyzed samples that were incorrectly classified by the interpretable CNN. We used VGG-16 networks for the binary classification of an animal category in the VOC Part dataset~\cite{SemanticPart}. We annotated the object-part name corresponding to each interpretable filter in the top interpretable layer. For each false positive sample without the target category, Fig.~\ref{fig:miscls} localized the image regions that were incorrectly detected as specific object parts by interpretable filters. This figure helped people understand the reason for misclassification.

\begin{table}[t]
\centering
\resizebox{\linewidth}{!}{\begin{tabular}{c|c|cc}
\multicolumn{4}{c}{Multi-category classification}\\
\hline
& {ILSVRC Part} & \multicolumn{2}{|c}{VOC Part}\\
\hline
& {logistic\footnotemark[3]}& {logistic\footnotemark[3]}& {softmax}\\
\hline
VGG-M & 96.73 & 93.88 & 81.93
\\
{interpretable} & {\bf97.99} & {\bf96.19} & {\bf88.03}
\\
\hline
VGG-S & 96.98 & 94.05 & 78.15
\\
{interpretable} & {\bf98.72} & {\bf96.78} & {\bf86.13}
\\
\hline
VGG-16 & -- & 97.97 & 89.71
\\
{interpretable} & -- & {\bf98.50} & {\bf91.60}
\\
\hline
\end{tabular}}
\resizebox{\linewidth}{!}{\begin{tabular}{c|ccc}
\multicolumn{4}{c}{Binary classification of a single category\textcolor{white}{\Large A}}\\
\hline
& {\scriptsize ILSVRC Part} & {\scriptsize VOC Part} & {\scriptsize CUB200}\\
\hline
AlexNet & {\bf96.28} & {\bf95.40} & {\bf95.59}
\\
interpretable w/o filter loss & -- & 93.98 & --\\
{interpretable} & 95.38 & 93.93 & 95.35
\\
\hline
VGG-M & {\bf97.34} & {\bf96.82} & {\bf97.34}
\\
interpretable w/o filter loss & -- & 93.13 & --\\
{interpretable} & 95.77 & 94.17 & 96.03
\\
\hline
VGG-S & {\bf97.62} & {\bf97.74} & {\bf97.24}
\\
interpretable w/o filter loss & -- & 93.83 & --\\
{interpretable} & 95.64 & 95.47 & 95.82
\\
\hline
VGG-16 & {\bf98.58} & {\bf98.66} & {\bf98.91}
\\
interpretable w/o filter loss & -- & 97.02 & --\\
{interpretable} & 96.67 & 95.39 & 96.51
\\
\hline
\end{tabular}}
\vspace{1pt}
\caption{Classification accuracy based on different datasets. In the binary classification of a single category, ordinary CNNs performed better, while in multi-category classification, interpretable CNNs exhibited superior performance.}
\label{tab:classification}
\end{table}

\begin{table}[t]
\centering
\resizebox{\linewidth}{!}{\begin{tabular}{c|cc}
& filters in & filters in\\
& ordinary CNNs & interpretable CNNs\\
\hline
AlexNet & 68.7 & {\bf75.1}\\
VGG-M & 69.9 & {\bf80.2}\\
VGG-16 & 72.1 & {\bf82.4}\\
\end{tabular}}
\caption{Classification accuracy based on a single filter. We reported the average accuracy to demonstrate the discrimination power of individual filters.}
\label{tab:filterClassify}
\end{table}

\begin{table}[t]
\centering
\resizebox{\linewidth}{!}{\begin{tabular}{c|c|c|c|c|c|c}
& bird & cow & cat & dog & horse & sheep\\
\hline
head & 19.2 & -- & -- & -- & 15.4 & --\\
neck & 21.2 & -- & -- & 5.8 & -- & --\\
torso & 36.5 & 32.7 & 3.8 & 38.5 & 75.0 & 52.0\\
hip \& tail & 5.8 & -- & -- & -- & -- & --\\
foot & 5.8 & -- & -- & -- & 9.6 & 3.8\\
wing & 11.5 & -- & -- & -- & -- & --\\
eye & -- & 30.8 & 55.8 & 11.5 & 7.7 & --\\
nose \& mouth & -- & 15.4 & 32.7 & 3.8 & 19.2 & --\\
side face & -- & 9.6 & -- & -- & -- & --\\
leg & -- & 11.5 & 5.8 & 23.1 & -- & --\\
ear \& horn & -- & -- & 1.9 & 17.3 & 17.3 & --\\
\end{tabular}}
\caption{Statistics of semantic meanings of interpretable filters. ``--'' indicates that the part is not selected as a label to describe the filter in a CNN. Except for CNNs for the bird and the horse, CNNs for other animals paid attention to detailed structures of the head. Thus, we annotated fine-grained parts inside the head for these CNNs.}
\label{tab:semantics}
\end{table}

% purity of voc
\begin{table*}[t]
\centering
\resizebox{0.7\linewidth}{!}{\begin{tabular}{l|cccccc|c}
\hline
& bird & cat & cow & dog & horse & sheep & {\bf Avg.}\\
\hline
VGG-16, interpretable &
{\bf0.568}&
{\bf0.656}&
{\bf0.534}&
{\bf0.573}&
{\bf0.570}&
{\bf0.492}&
{\bf0.566}
\\
VGG-16 + mask layer, w/o filter loss &
{0.385}&
{0.373}&
{0.342}&
{0.435}&
{0.382}&
{0.303}&
{0.370}\\
\hline
VGG-M, interpretable &
{\bf0.444}&
{\bf0.616}&
{\bf0.395}&
{\bf0.540}&
{\bf0.408}&
{\bf0.387}&
{\bf0.465}
\\
VGG-M + mask layer, w/o filter loss &
{0.230}&
{0.341}&
{0.185}&
{0.295}&
{0.250}&
{0.215}&
{0.253}\\
\hline
VGG-S, interpretable &
{\bf0.418}&
{\bf0.437}&
{\bf0.390}&
{\bf0.398}&
{\bf0.421}&
{\bf0.369}&
{\bf0.406}
\\
VGG-S + mask layer, w/o filter loss &
{0.224}&
{0.312}&
{0.165}&
{0.234}&
{0.208}&
{0.161}&
{0.217}\\
\hline
\end{tabular}}
\vspace{1pt}
\caption{Semantic purity of neural activations of interpretable filters learned with and without the filter loss from the VOC Part dataset. Filters learned with the filter loss exhibited significantly higher semantic purity than those learned without filter loss.}
\label{tab:concentration_voc}
\end{table*}

% purity of imagenet
\begin{table*}[t]
  \centering
  \resizebox{\linewidth}{!}{\begin{tabular}{p{4.5cm}|ccccccccccc}
    \hline
    \!\!\!&\!\! gold. \!\!\!&\!\! bird \!\!\!&\!\! frog \!\!\!&\!\! turt. \!\!\!&\!\! liza. \!\!\!&\!\! koala \!\!\!&\!\! lobs. \!\!\!&\!\! dog \!\!\!&\!\! fox \!\!\!&\!\! cat \!\!\!&\!\! lion\\
    \!\!\! {\footnotesize VGG-16, interpretable}\!\!\!&\!\!
    {\bf0.614}\!\!\!&\!\!
    {\bf0.624}\!\!\!&\!\!
    {\bf0.585}\!\!\!&\!\!
    {\bf0.527}\!\!\!&\!\!
    {\bf0.531}\!\!\!&\!\!
    {\bf0.607}\!\!\!&\!\!
    {\bf0.581}\!\!\!&\!\!
    {\bf0.606}\!\!\!&\!\!
    {\bf0.660}\!\!\!&\!\!
    {\bf0.609}\!\!\!&\!\!
    {\bf0.572}
    \\
    \!\!\! {\footnotesize VGG-16 + mask layer, w/o filter loss}\!\!\!&\!\!
    {0.331}\!\!\!&\!\!
    {0.441}\!\!\!&\!\!
    {0.342}\!\!\!&\!\!
    {0.352}\!\!\!&\!\!
    {0.362}\!\!\!&\!\!
    {0.339}\!\!\!&\!\!
    {0.344}\!\!\!&\!\!
    {0.482}\!\!\!&\!\!
    {0.449}\!\!\!&\!\!
    {0.393}\!\!\!&\!\!
    {0.344}
    \\
    \cline{1-1}

    \!\!\! {\footnotesize VGG-M, interpretable}\!\!\!&\!\!
    {\bf0.533}\!\!\!&\!\!
    {\bf0.442}\!\!\!&\!\!
    {\bf0.385}\!\!\!&\!\!
    {\bf0.444}\!\!\!&\!\!
    {\bf0.430}\!\!\!&\!\!
    {\bf0.408}\!\!\!&\!\!
    {\bf0.430}\!\!\!&\!\!
    {\bf0.587}\!\!\!&\!\!
    {\bf0.564}\!\!\!&\!\!
    {\bf0.466}\!\!\!&\!\!
    {\bf0.572}
    \\
    \!\!\! {\footnotesize VGG-M + mask layer, w/o filter loss}\!\!\!&\!\!
    {0.307}\!\!\!&\!\!
    {0.251}\!\!\!&\!\!
    {0.249}\!\!\!&\!\!
    {0.198}\!\!\!&\!\!
    {0.266}\!\!\!&\!\!
    {0.255}\!\!\!&\!\!
    {0.212}\!\!\!&\!\!
    {0.384}\!\!\!&\!\!
    {0.393}\!\!\!&\!\!
    {0.389}\!\!\!&\!\!
    {0.293}
    \\
    \cline{1-1}

    \!\!\! {\footnotesize VGG-S, interpretable}\!\!\!&\!\!
    {\bf0.454}\!\!\!&\!\!
    {\bf0.453}\!\!\!&\!\!
    {\bf0.389}\!\!\!&\!\!
    {\bf0.409}\!\!\!&\!\!
    {\bf0.389}\!\!\!&\!\!
    {\bf0.414}\!\!\!&\!\!
    {\bf0.398}\!\!\!&\!\!
    {\bf0.431}\!\!\!&\!\!
    {\bf0.456}\!\!\!&\!\!
    {\bf0.442}\!\!\!&\!\!
    {\bf0.395}
    \\
    \!\!\! {\footnotesize VGG-S + mask layer, w/o filter loss}\!\!\!&\!\!
    {0.245}\!\!\!&\!\!
    {0.265}\!\!\!&\!\!
    {0.186}\!\!\!&\!\!
    {0.189}\!\!\!&\!\!
    {0.216}\!\!\!&\!\!
    {0.212}\!\!\!&\!\!
    {0.209}\!\!\!&\!\!
    {0.316}\!\!\!&\!\!
    {0.330}\!\!\!&\!\!
    {0.284}\!\!\!&\!\!
    {0.197}
    \\

    \hline
    \!\!\!&\!\! tiger \!\!\!&\!\! bear \!\!\!&\!\! rabb. \!\!\!&\!\! hams. \!\!\!&\!\! squi. \!\!\!&\!\! horse \!\!\!&\!\! zebra \!\!\!&\!\! swine \!\!\!&\!\! hippo. \!\!\!&\!\! catt. \!\!\!&\!\! sheep\\

    \!\!\! {\footnotesize VGG-16, interpretable}\!\!\!&\!\!
    {\bf0.583}\!\!\!&\!\!
    {\bf0.614}\!\!\!&\!\!
    {\bf0.535}\!\!\!&\!\!
    {\bf0.679}\!\!\!&\!\!
    {\bf0.677}\!\!\!&\!\!
    {\bf0.559}\!\!\!&\!\!
    {\bf0.532}\!\!\!&\!\!
    {\bf0.530}\!\!\!&\!\!
    {\bf0.564}\!\!\!&\!\!
    {\bf0.466}\!\!\!&\!\!
    {\bf0.572}
    \\

    \!\!\! {\footnotesize VGG-16 + mask layer, w/o filter loss}\!\!\!&\!\!
    {0.368}\!\!\!&\!\!
    {0.369}\!\!\!&\!\!
    {0.364}\!\!\!&\!\!
    {0.356}\!\!\!&\!\!
    {0.419}\!\!\!&\!\!
    {0.407}\!\!\!&\!\!
    {0.343}\!\!\!&\!\!
    {0.363}\!\!\!&\!\!
    {0.393}\!\!\!&\!\!
    {0.389}\!\!\!&\!\!
    {0.293}
    \\
    \cline{1-1}

    \!\!\! {\footnotesize VGG-M, interpretable}\!\!\!&\!\!
    {\bf0.455}\!\!\!&\!\!
    {\bf0.442}\!\!\!&\!\!
    {\bf0.472}\!\!\!&\!\!
    {\bf0.452}\!\!\!&\!\!
    {\bf0.434}\!\!\!&\!\!
    {\bf0.397}\!\!\!&\!\!
    {\bf0.421}\!\!\!&\!\!
    {\bf0.412}\!\!\!&\!\!
    {\bf0.420}\!\!\!&\!\!
    {\bf0.425}\!\!\!&\!\!
    {\bf0.420}
    \\

    \!\!\! {\footnotesize VGG-M + mask layer, w/o filter loss}\!\!\!&\!\!
    {0.345}\!\!\!&\!\!
    {0.345}\!\!\!&\!\!
    {0.316}\!\!\!&\!\!
    {0.245}\!\!\!&\!\!
    {0.320}\!\!\!&\!\!
    {0.283}\!\!\!&\!\!
    {0.224}\!\!\!&\!\!
    {0.278}\!\!\!&\!\!
    {0.290}\!\!\!&\!\!
    {0.397}\!\!\!&\!\!
    {0.284}
    \\
    \cline{1-1}

    \!\!\! {\footnotesize VGG-S, interpretable}\!\!\!&\!\!
    {\bf0.448}\!\!\!&\!\!
    {\bf0.419}\!\!\!&\!\!
    {\bf0.411}\!\!\!&\!\!
    {\bf0.405}\!\!\!&\!\!
    {\bf0.416}\!\!\!&\!\!
    {\bf0.430}\!\!\!&\!\!
    {\bf0.469}\!\!\!&\!\!
    {\bf0.384}\!\!\!&\!\!
    {\bf0.403}\!\!\!&\!\!
    {\bf0.439}\!\!\!&\!\!
    {\bf0.404}
    \\

    \!\!\! {\footnotesize VGG-S + mask layer, w/o filter loss}\!\!\!&\!\!
    {0.215}\!\!\!&\!\!
    {0.197}\!\!\!&\!\!
    {0.212}\!\!\!&\!\!
    {0.179}\!\!\!&\!\!
    {0.217}\!\!\!&\!\!
    {0.210}\!\!\!&\!\!
    {0.215}\!\!\!&\!\!
    {0.177}\!\!\!&\!\!
    {0.200}\!\!\!&\!\!
    {0.323}\!\!\!&\!\!
    {0.196}
    \\
    \hline
    \!\!\!&\!\! ante. \!\!\!&\!\! camel \!\!\!&\!\! otter \!\!\!&\!\! arma. \!\!\!&\!\! monk. \!\!\!&\!\! elep. \!\!\!&\!\! red pa. \!\!\!&\!\! gia.pa. \!\!\!&\!\! \!\!\!&\!\! \!\!\!&\!\! {\bf Avg.}\\

    \!\!\! {\footnotesize VGG-16, interpretable}\!\!\!&\!\!
    {\bf0.573}\!\!\!&\!\!
    {\bf0.544}\!\!\!&\!\!
    {\bf0.565}\!\!\!&\!\!
    {\bf0.855}\!\!\!&\!\!
    {\bf0.657}\!\!\!&\!\!
    {\bf0.562}\!\!\!&\!\!
    {\bf0.718}\!\!\!&\!\!
    {\bf0.697}\!\!\!&\!\!
    \!\!\!&\!\!
    \!\!\!&\!\!
    {\bf0.595}
    \\

    \!\!\! {\footnotesize VGG-16 + mask layer, w/o filter loss}\!\!\!&\!\!
    {0.440}\!\!\!&\!\!
    {0.367}\!\!\!&\!\!
    {0.315}\!\!\!&\!\!
    {0.321}\!\!\!&\!\!
    {0.377}\!\!\!&\!\!
    {0.333}\!\!\!&\!\!
    {0.391}\!\!\!&\!\!
    {0.383}\!\!\!&\!\!
    \!\!\!&\!\!
    \!\!\!&\!\!
    {0.372}
    \\
    \cline{1-1}
    \!\!\! {\footnotesize VGG-M, interpretable}\!\!\!&\!\!
    {\bf0.584}\!\!\!&\!\!
    {\bf0.435}\!\!\!&\!\!
    {\bf0.441}\!\!\!&\!\!
    {\bf0.419}\!\!\!&\!\!
    {\bf0.400}\!\!\!&\!\!
    {\bf0.399}\!\!\!&\!\!
    {\bf0.541}\!\!\!&\!\!
    {\bf0.468}\!\!\!&\!\!
    \!\!\!&\!\!
    \!\!\!&\!\!
    {\bf0.459}
    \\

    \!\!\! {\footnotesize VGG-M + mask layer, w/o filter loss}\!\!\!&\!\!
    {0.411}\!\!\!&\!\!
    {0.315}\!\!\!&\!\!
    {0.232}\!\!\!&\!\!
    {0.175}\!\!\!&\!\!
    {0.225}\!\!\!&\!\!
    {0.173}\!\!\!&\!\!
    {0.334}\!\!\!&\!\!
    {0.343}\!\!\!&\!\!
    \!\!\!&\!\!
    \!\!\!&\!\!
    {0.285}
    \\
    \cline{1-1}

    \!\!\! {\footnotesize VGG-S, interpretable}\!\!\!&\!\!
    {\bf0.471}\!\!\!&\!\!
    {\bf0.386}\!\!\!&\!\!
    {\bf0.386}\!\!\!&\!\!
    {\bf0.436}\!\!\!&\!\!
    {\bf0.394}\!\!\!&\!\!
    {\bf0.408}\!\!\!&\!\!
    {\bf0.459}\!\!\!&\!\!
    {\bf0.471}\!\!\!&\!\!
    \!\!\!&\!\!
    \!\!\!&\!\!
    {\bf0.420}
    \\
    \!\!\! {\footnotesize VGG-S + mask layer, w/o filter loss}\!\!\!&\!\!
    {0.388}\!\!\!&\!\!
    {0.202}\!\!\!&\!\!
    {0.184}\!\!\!&\!\!
    {0.183}\!\!\!&\!\!
    {0.188}\!\!\!&\!\!
    {0.185}\!\!\!&\!\!
    {0.262}\!\!\!&\!\!
    {0.252}\!\!\!&\!\!
    \!\!\!&\!\!
    \!\!\!&\!\!
    {0.226}
    \\
    \hline
  \end{tabular}}
  \vspace{1pt}
  \caption{Semantic purity of neural activations of interpretable filters learned the ILSVRC 2013 DET Animal-Part dataset with and without the filter loss. Filters learned with the filter loss exhibited significantly higher semantic purity than those learned without the filter loss.}
  \label{tab:concentration_imagenet}
  \end{table*}

% purity visulization
\begin{figure*}[t]
\centering
\includegraphics[width=\linewidth]{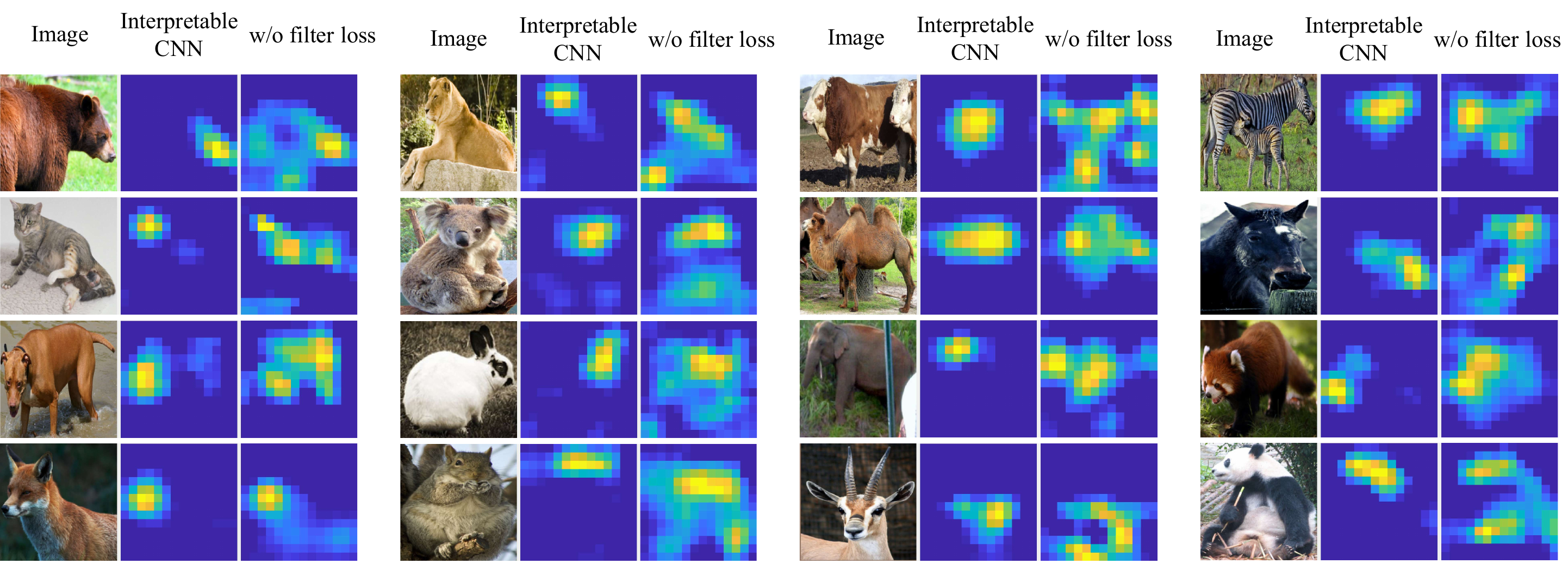}
\caption{Visualization of neural activations (before the mask layer) of interpretable filters learned with and without the filter loss. We visualized neural activations of the first interpretable conv-layer before the mask layer in the CNN. In comparison, visualization results in Fig.~\ref{fig:visual} correspond to feature maps after the mask layer. Filters trained with the filter loss tended to generate more concentrated neural activations and have higher semantic purity that filters learned without the filter loss.}
\label{fig:purity}
\end{figure*}

% instability of voc
\begin{table*}[t]
  \centering
  \resizebox{0.7\linewidth}{!}{\begin{tabular}{l|cccccc|c}
  \hline
  & bird & cat & cow & dog & horse & sheep & {\bf Avg.}\\
  \hline
  VGG-16 &
  {0.144}&
  {0.134}&
  {0.146}&
  {0.127}&
  {0.141}&
  {0.142}&
  {0.139}\\
  VGG-16 + mask layer, w/o filter loss &
  {0.141}&
  {\bf0.126}&
  {0.139}&
  {0.124}&
  {0.138}&
  {0.136}&
  {0.134}\\
  VGG-16, interpretable &
  {\bf0.130}&
  {0.127}&
  {\bf0.134}&
  {\bf0.109}&
  {\bf0.131}&
  {\bf0.126}&
  {\bf0.126}
  \\
  \hline
  \end{tabular}}
  \vspace{1pt}
  \caption{Location instability of filters ($\mathbb{E}_{f,k}[D_{f,k}]$) in the first conv-layer. The CNN was trained for the binary classification of a single category using the VOC Part dataset~\cite{SemanticPart}. For the baseline, we added a conv-layer to the ordinary VGG-16 network, and selected the corresponding conv-layer in the network to enable fair comparisons. Interpretable filters learned with both the filter loss and the mask layer exhibited much lower localization instability than those learned with the mask layer but without the filter loss.}
  \label{tab:instability_before_mask_voc}
  \end{table*}

  \begin{table*}[t]
    \centering
    \resizebox{\linewidth}{!}{\begin{tabular}{p{4.5cm}|ccccccccccc}
      \hline
      \!\!\!&\!\! gold. \!\!\!&\!\! bird \!\!\!&\!\! frog \!\!\!&\!\! turt. \!\!\!&\!\! liza. \!\!\!&\!\! koala \!\!\!&\!\! lobs. \!\!\!&\!\! dog \!\!\!&\!\! fox \!\!\!&\!\! cat \!\!\!&\!\! lion\\
      \!\!\! {\footnotesize VGG-16}\!\!\!&\!\!
      {0.152}\!\!\!&\!\!
      {0.152}\!\!\!&\!\!
      {0.142}\!\!\!&\!\!
      {0.150}\!\!\!&\!\!
      {0.167}\!\!\!&\!\!
      {0.125}\!\!\!&\!\!
      {0.126}\!\!\!&\!\!
      {0.139}\!\!\!&\!\!
      {0.137}\!\!\!&\!\!
      {0.149}\!\!\!&\!\!
      {0.134}
      \\
      \!\!\! {\footnotesize VGG-16 + mask layer, w/o filter loss}\!\!\!&\!\!
      {0.143}\!\!\!&\!\!
      {0.148}\!\!\!&\!\!
      {0.141}\!\!\!&\!\!
      {0.146}\!\!\!&\!\!
      {0.162}\!\!\!&\!\!
      {0.121}\!\!\!&\!\!
      {0.120}\!\!\!&\!\!
      {0.137}\!\!\!&\!\!
      {0.130}\!\!\!&\!\!
      {0.142}\!\!\!&\!\!
      {0.130}
      \\
      \!\!\! {\footnotesize VGG-16, interpretable}\!\!\!&\!\!
      {\bf0.105}\!\!\!&\!\!
      {\bf0.127}\!\!\!&\!\!
      {\bf0.131}\!\!\!&\!\!
      {\bf0.139}\!\!\!&\!\!
      {\bf0.157}\!\!\!&\!\!
      {\bf0.102}\!\!\!&\!\!
      {\bf0.118}\!\!\!&\!\!
      {\bf0.123}\!\!\!&\!\!
      {\bf0.105}\!\!\!&\!\!
      {\bf0.129}\!\!\!&\!\!
      {\bf0.106}
      \\
      \cline{1-1}
      \hline
      \!\!\!&\!\! tiger \!\!\!&\!\! bear \!\!\!&\!\! rabb. \!\!\!&\!\! hams. \!\!\!&\!\! squi. \!\!\!&\!\! horse \!\!\!&\!\! zebra \!\!\!&\!\! swine \!\!\!&\!\! hippo. \!\!\!&\!\! catt. \!\!\!&\!\! sheep\\

      \!\!\! {\footnotesize VGG-16}\!\!\!&\!\!
      {0.143}\!\!\!&\!\!
      {0.139}\!\!\!&\!\!
      {0.143}\!\!\!&\!\!
      {0.124}\!\!\!&\!\!
      {0.145}\!\!\!&\!\!
      {0.147}\!\!\!&\!\!
      {0.155}\!\!\!&\!\!
      {0.138}\!\!\!&\!\!
      {0.139}\!\!\!&\!\!
      {0.142}\!\!\!&\!\!
      {0.148}
      \\
      \!\!\! {\footnotesize VGG-16 + mask layer, w/o filter loss}\!\!\!&\!\!
      {0.137}\!\!\!&\!\!
      {0.131}\!\!\!&\!\!
      {0.140}\!\!\!&\!\!
      {0.119}\!\!\!&\!\!
      {0.134}\!\!\!&\!\!
      {0.143}\!\!\!&\!\!
      {0.142}\!\!\!&\!\!
      {0.133}\!\!\!&\!\!
      {0.131}\!\!\!&\!\!
      {0.134}\!\!\!&\!\!
      {0.145}
      \\
      \!\!\! {\footnotesize VGG-16, interpretable}\!\!\!&\!\!
      {\bf0.112}\!\!\!&\!\!
      {\bf0.118}\!\!\!&\!\!
      {\bf0.122}\!\!\!&\!\!
      {\bf0.098}\!\!\!&\!\!
      {\bf0.106}\!\!\!&\!\!
      {\bf0.136}\!\!\!&\!\!
      {\bf0.095}\!\!\!&\!\!
      {\bf0.125}\!\!\!&\!\!
      {\bf0.120}\!\!\!&\!\!
      {\bf0.126}\!\!\!&\!\!
      {\bf0.128}
      \\
      \cline{1-1}
      \hline
      \!\!\!&\!\! ante. \!\!\!&\!\! camel \!\!\!&\!\! otter \!\!\!&\!\! arma. \!\!\!&\!\! monk. \!\!\!&\!\! elep. \!\!\!&\!\! red pa. \!\!\!&\!\! gia.pa. \!\!\!&\!\! \!\!\!&\!\! \!\!\!&\!\! {\bf Avg.}\\

      \!\!\! {\footnotesize VGG-16}\!\!\!&\!\!
      {0.141}\!\!\!&\!\!
      {0.144}\!\!\!&\!\!
      {0.155}\!\!\!&\!\!
      {0.156}\!\!\!&\!\!
      {0.134}\!\!\!&\!\!
      {0.127}\!\!\!&\!\!
      {0.139}\!\!\!&\!\!
      {0.126}\!\!\!&\!\!
      \!\!\!&\!\!
      \!\!\!&\!\!
      {0.142}
      \\
      \!\!\! {\footnotesize VGG-16 + mask layer, w/o filter loss}\!\!\!&\!\!
      {0.134}\!\!\!&\!\!
      {0.142}\!\!\!&\!\!
      {0.147}\!\!\!&\!\!
      {0.153}\!\!\!&\!\!
      {0.129}\!\!\!&\!\!
      {0.124}\!\!\!&\!\!
      {0.125}\!\!\!&\!\!
      {0.116}\!\!\!&\!\!
      \!\!\!&\!\!
      \!\!\!&\!\!
      {0.136}
      \\
      \!\!\! {\footnotesize VGG-16, interpretable}\!\!\!&\!\!
      {\bf0.119}\!\!\!&\!\!
      {\bf0.131}\!\!\!&\!\!
      {\bf0.134}\!\!\!&\!\!
      {\bf0.118}\!\!\!&\!\!
      {\bf0.112}\!\!\!&\!\!
      {\bf0.109}\!\!\!&\!\!
      {\bf0.098}\!\!\!&\!\!
      {\bf0.088}\!\!\!&\!\!
      \!\!\!&\!\!
      \!\!\!&\!\!
      {\bf0.118}
      \\
      \cline{1-1}
      \hline
    \end{tabular}}
    \vspace{1pt}
    \caption{Location instability of filters ($\mathbb{E}_{f,k}[D_{f,k}]$) in the first interpretable conv-layer. The CNN was trained for the binary classification of a single category using the ILSVRC 2013 DET Animal-Part dataset~\cite{CNNAoG}. For the baseline, we added a conv-layer to the ordinary VGG-16 network, and selected the corresponding conv-layer in the network to enable fair comparisons. Interpretable filters learned with both the filter loss and the mask layer exhibited much lower localization instability than those learned with the mask layer but without the filter loss.}
    \label{tab:instability_before_mask_imagenet}
    \end{table*}

    \begin{table*}[t]
      \centering
      \resizebox{0.65\linewidth}{!}{\begin{tabular}{c|cc|cc}
      \hline
      Model & \multicolumn{2}{|c|}{Logistic log loss} & \multicolumn{2}{|c}{Softmax log loss}\\
      \hline
      & on target & on other & on target & on other\\
      & categories & categories & categories & categories\\
      \hline
      VGG-M, interpretable & $107.9$ & $7.5$ & $6.2$ & $1.1$\\
      {\small w/o filter loss} & $18.6$ & $6.3$ & $4.7$ & $1.1$\\
      \hline
      VGG-S, interpretable & $32.1$ & $10.5$ & $18.8$ & $3.1$\\
      {\small w/o filter loss} & $18.9$ & $8.2$ & $11.5$ & $2.3$\\
      \hline
      VGG-16, interpretable & $948.9$ & $81.5$ & $106.7$ & $5.1$\\
      {\small w/o filter loss} & $40.5$ & $12.9$ & $67.0$ & $6.0$\\
      \hline
      \end{tabular}}
      \caption{The mean value of neural activations on the target categories and those on other categories. Filters learned with the filter loss exhibited were usually more discriminative than those learned without the filter loss.}
      \label{tab:activation}
      \end{table*}

\begin{table}[t]
\centering
\begin{tabular}{c|cc}
\hline
Model & Original CNN & Interpretable CNN\\
\hline
VGG-M & 0.00302$\pm$0.00123 & 0.00243$\pm$0.00120\\
VGG-S & 0.00305$\pm$0.00120 & 0.00266$\pm$0.00133\\
VGG-16 & 0.00293$\pm$0.00128 & 0.00280$\pm$0.00152\\
\hline
\end{tabular}
\caption{Average adversarial distortion of the original CNN and the interpretable CNN.}
\label{tab:adversarial}
\end{table}

\subsection{Effects of the filter loss}

In this section, we evaluated effects of the filter loss. We compared the interpretable CNN learned with the filter loss with that without the filter loss (\emph{i.e.} only using the mask layer without the filter loss).

\subsubsection{Semantic purity of neural activations}

We proposed a metric to measure the semantic purity of neural activations of a filter. If a filter was activated at multiple locations besides the highest peak (\emph{i.e.} the one corresponding to the target part), we considered this filter to have low semantic purity.

The semantic purity of a filter was measured as the ratio of neural activations within the range of the mask to all neural activations of the filter. In other words, the purity of a filter indicated that whether a filter was learned to represent a single part or represent multiple parts.

Let $x\in \mathbb{R}^{n\times n}$ be neural activations of a filter (after the ReLU layer and before the mask layer). The corresponding mask was given as $T_{\hat{\mu}}\in\mathbb{R}^{n\times n}$. The purity of neural activations was defined as $purity=\frac{\sum_{f}\sum_{i,j}\max(0,x_{ij})\cdot\textbf{1}(T_{\hat{\mu},ij}>0)}{\sum_{f}\sum_{i,j} \max(0,x_{ij})}$. $\textbf{1}(\cdot)$ was the indicator function, which returns $1$ if the condition in the braces was satisfied, and returns 0 otherwise.  The purity was supposed to be higher if neural activations were more concentrated.

We compared the purity of neural activations between the interpretable CNN and the CNN trained with the mask layer but without filter loss. We constructed these CNNs based on the architectures of VGG-16, VGG-M and VGG-S, and learned the CNNs for binary classification on an animal category in the VOC Part dataset and the ILSVRC 2013 DET Animal-Part dataset. Experimental results are shown in Table~\ref{tab:concentration_voc} and Table~\ref{tab:concentration_imagenet}. It demonstrated that filters learned with the filter loss exhibited higher semantic purity than those learned without the filter loss. The filter loss forced each filter to exclusively represented a single object part during the training process.

\subsubsection{Visualization of filters}

Besides the quantitative analysis of neural activation purity, we visualized filter activations to compare the interpretable CNN and the CNN trained without the filter loss. Fig.~\ref{fig:purity} visualized neural activations of the filter in the first interpretable conv-layer of VGG-16 before the mask layer. Visualization results demonstrated that filters trained with the filter loss could generate more concentrated neural activations.

\subsubsection{Location instability}

We compared the location instability among interpretable filters learned with the filter loss, those learned without the filter loss, and ordinary filters. Here, we used filers in the first interpretable conv-layer of the interpretable CNN and filters in the corresponding conv-layer of the traditional CNNs for comparison.

We constructed the competing CNNs based on the VGG-16 architecture, and these CNNs were trained for single-category classification based on the VOC Part dataset and the ILSVRC 2013 DET Animal-Part dataset. As shown in Table~\ref{tab:instability_before_mask_voc} and Table~\ref{tab:instability_before_mask_imagenet}, the filter loss forced each filter to focus on a specific object part and reduced the location instability.

\subsubsection{Activation magnitudes}

We further tested effects of the interpretable loss on neural activations among different categories. We used the VGG-M, VGG-S, and VGG-16 networks with either the logistic log loss or the softmax loss, which was trained to classify animal categories in the VOC Part dataset~\cite{SemanticPart}. For each interpretable filter, given images of its target category, we recorded their neural activations (\emph{i.e.} recording the maximal activation value in each of their feature maps). At the same time, we also recorded neural activations on other categories of the filter. The interpretable filter was supposed to activate much more strongly on its target category than on other (unrelated) categories. We collected all activation records on corresponding categories of all filters, and their mean value is reported in Table~\ref{tab:activation}. In comparison, we also computed the mean value of all activations on unrelated categories of all filters in Table~\ref{tab:activation}. This table shows that the interpretable filter was usually activated more strongly on the target category than on other categories. Furthermore, we also compared the proposed interpretable CNN with the ablation baseline \textit{w/o filter loss}, in which the CNN was learned without the filter loss. Table~\ref{tab:activation} shows that the filter loss made each filter more prone to being triggered by a single category, \emph{i.e.} boosting the feature interpretability.

\section{Conclusion and discussions}

In this paper, we have proposed a general method to enhance feature interpretability of CNNs. We design a loss to push a filter in high conv-layers towards the representation of an object part during the learning process without any part annotations. Experiments have shown that each interpretable filter consistently represents a certain object part of a category through different input images. In comparison, each filter in the traditional CNN usually represents a mixture of parts and textures.

Meanwhile, the interpretable CNN still has some drawbacks. First, in the scenario of multi-category classification, filters in a conv-layer are assigned with different categories. In this way, when we need to classify a large number of categories, theoretically, each category can only obtain a few filters, which will decrease a bit the classification performance. Otherwise, the interpretable conv-layer must contain lots of filters to enable the classification of a large number of categories. Second, the learning of the interpretable CNN has a strong assumption, \emph{i.e.} each input image must contain a single object, which limits the applicability of the interpretable CNN. Third, the filter loss is only suitable to learn high conv-layers, because low conv-layers usually represent textures, instead of object parts. Finally, the interpretable CNN is not suitable to encode textural patterns.

\ifCLASSOPTIONcompsoc
  \section*{Acknowledgments}
\else
  \section*{Acknowledgment}
\fi

This work is partially supported by National Natural Science Foundation of China (U19B2043 and 61906120), DARPA XAI Award N66001-17-2-4029, NSF IIS 1423305, and ARO project W911NF1810296.

\appendix
\section*{Proof of Equation~\eqref{eqn:grad}}
\begin{small}
\begin{eqnarray}
\frac{\partial{\bf Loss}}{\partial x_{ij}}\!&\!=\!&\!-\sum_{\mu\in{\boldsymbol\Omega}}p(\mu)\Big\{\frac{\partial p(x|\mu)}{\partial x_{ij}}\big[\log p(x|\mu)-\log p(x)+1\big]\nonumber\\
&&-p(x|\mu)\frac{\partial\log p(x)}{\partial x_{ij}}\Big\}\\
\!&\!=\!&\!-\sum_{\mu\in{\boldsymbol\Omega}}p(\mu)\Big\{\frac{\partial p(x|\mu)}{\partial x_{ij}}\big[\log p(x|\mu)-\log p(x)+1\big]\nonumber\\
&&-p(x|\mu)\frac{1}{p(x)}\frac{\partial p(x)}{\partial x_{ij}}\Big\}\nonumber\\
\!&\!=\!&\!-\sum_{\mu\in{\boldsymbol\Omega}}p(\mu)\Big\{\frac{\partial p(x|\mu)}{\partial x_{ij}}\big[\log p(x|\mu)-\log p(x)+1\big]\nonumber\\
&&-p(x|\mu)\frac{1}{p(x)}\sum_{\mu'}\Big[p(\mu')\frac{\partial p(x|\mu')}{\partial x_{ij}}\Big]\Big\}\nonumber\\
\!&\!=\!&\!-\sum_{\mu\in{\boldsymbol\Omega}}p(\mu)\Big\{\frac{\partial p(x|\mu)}{\partial x_{ij}}\big[\log p(x|\mu)-\log p(x)+1\big]\Big\}\nonumber\\
&&+\sum_{\mu\in{\boldsymbol\Omega}}p(\mu)\frac{\partial p(x|\mu)}{\partial x_{ij}}\frac{\sum_{\mu'}p(\mu')p(x|\mu')}{p(x)}\nonumber\\
\!&\!=\!&\!-\sum_{\mu\in{\boldsymbol\Omega}}p(\mu)\Big\{\frac{\partial p(x|\mu)}{\partial x_{ij}}\big[\log p(x|\mu)-\log p(x)+1\big]\Big\}\nonumber\\
&&+\sum_{\mu\in{\boldsymbol\Omega}}p(\mu)\frac{\partial p(x|\mu)}{\partial x_{ij}}\nonumber\\
\!&\!=\!&\!-\sum_{\mu\in{\boldsymbol\Omega}}\frac{\partial p(x|\mu)}{\partial x_{ij}}p(\mu)\big[\log p(x|\mu)-\log p(x)\big]\nonumber\\
\!&\!=\!&\!-\sum_{\mu\in{\boldsymbol\Omega}}\frac{t_{ij}p(\mu)e^{tr(x\cdot T)}}{Z_{\mu}}\Big\{tr(x\cdot T)-\log\big[Z_{\mu}p(x)\big]\Big\}\nonumber
\end{eqnarray}
\end{small}

\section*{Proof of Equation~\eqref{eqn:understand}}

\begin{small}
\begin{eqnarray}
&&\!\!\!\!\!\!\!{\bf Loss}=-MI({\bf X};{\boldsymbol\Omega})\qquad//\;\;{\boldsymbol\Omega}=\{\mu^{-},\mu_{1},\mu_{2},\ldots,\mu_{n^2}\}\nonumber\\
\!&\!=\!&\!-H({\boldsymbol\Omega})+H({\boldsymbol\Omega}|{\bf X})\\
\!&\!=\!&\!-H({\boldsymbol\Omega})-\sum_{x}p(x)\sum_{\mu\in{\boldsymbol\Omega}}p(\mu|x)\log p(\mu|x)\nonumber\\
\!&\!=\!&\!-H({\boldsymbol\Omega})-\sum_{x}p(x)\Big\{p(\mu^{-}|x)\log p(\mu^{-}|x)\nonumber\\
&&+\sum_{\mu\in{\boldsymbol\Omega}^{+}}p(\mu|x)\log p(\mu|x)\Big\}\nonumber\\
\!&\!=\!&\!-H({\boldsymbol\Omega})-\sum_{x}p(x)\Big\{p(\mu^{-}|x)\log p(\mu^{-}|x)\nonumber\\
&&+\sum_{\mu\in{\boldsymbol\Omega}^{+}}p(\mu|x)\log\big[\frac{p(\mu|x)}{p({\boldsymbol\Omega}^{+}|x)}p({\boldsymbol\Omega}^{+}|x)\big]\Big\}\nonumber\\
\!&\!=\!&\!-H({\boldsymbol\Omega})-\sum_{x}p(x)\Big\{p(\mu^{-}|x)\log p(\mu^{-}|x)\nonumber\\
&&+p({\boldsymbol\Omega}^{+}|x)\log p({\boldsymbol\Omega}^{+}|x)+\sum_{\mu\in{\boldsymbol\Omega}^{+}}p(\mu|x)\log\frac{p(\mu|x)}{p({\boldsymbol\Omega}^{+}|x)}\Big\}\nonumber\\
\!&\!=\!&\!-H({\boldsymbol\Omega})+H({\boldsymbol\Omega}'=\{\mu^{-},{\boldsymbol\Omega}^{+}\}|{\bf X})\nonumber\\
&&+\sum_{x}p({\boldsymbol\Omega}^{+},x)H({\boldsymbol\Omega}^{+}|X=x)\nonumber
\end{eqnarray}
\end{small}
where $p({\boldsymbol\Omega}^{+}|x)=\sum_{\mu\in{\boldsymbol\Omega}^{+}}p(\mu|x)$, $H({\boldsymbol\Omega}^{+}|X=x)=\sum_{\mu\in{\boldsymbol\Omega}^{+}}\tilde{p}(\mu|X=x)\log\tilde{p}(\mu|X=x)$, $\tilde{p}(\mu|X=x)=\frac{p(\mu|x)}{p({\boldsymbol\Omega}^{+}|x)}$.

\ifCLASSOPTIONcaptionsoff
  \newpage
\fi

\bibliographystyle{IEEE}
\bibliography{TheBib}

\vspace{-30pt}
\begin{IEEEbiography}[{\includegraphics[width=1in,height=1.25in,clip,keepaspectratio]{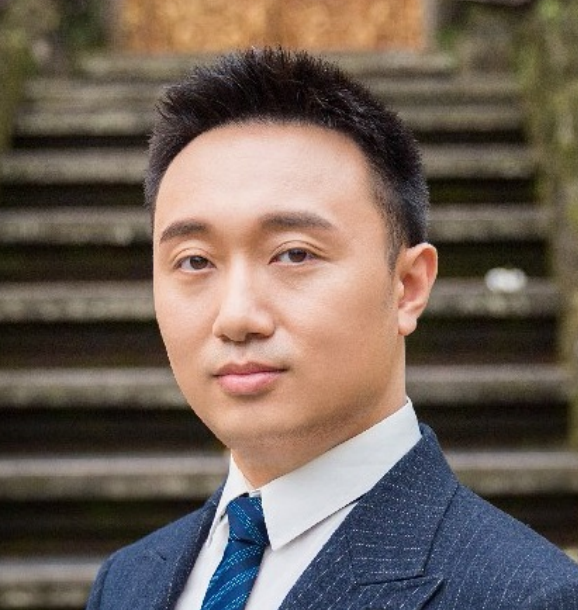}}]{Quanshi Zhang}
received the B.S. degree in machine intelligence from the Peking University, China, in 2009 and M.S. and Ph.D. degrees in center for spatial information science from the University of Tokyo, Japan, in 2011 and 2014, respectively. In 2014, he went to the University of California, Los Angeles, as a post-doctoral associate. Now, he is an associate professor at the Shanghai Jiao Tong University. His research interests include computer vision, machine learning, and robotics.
\end{IEEEbiography}

\vspace{-30pt}
\begin{IEEEbiography}[{\includegraphics[width=1in,height=1.25in,clip,keepaspectratio]{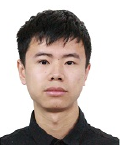}}]{Xin Wang} is a Ph.D. student an internship student at the Shanghai Jiao Tong University. His research mainly focuses on machine learning and computer vision.
\end{IEEEbiography}

\begin{IEEEbiography}[{\includegraphics[width=1in,height=1.25in,clip,keepaspectratio]{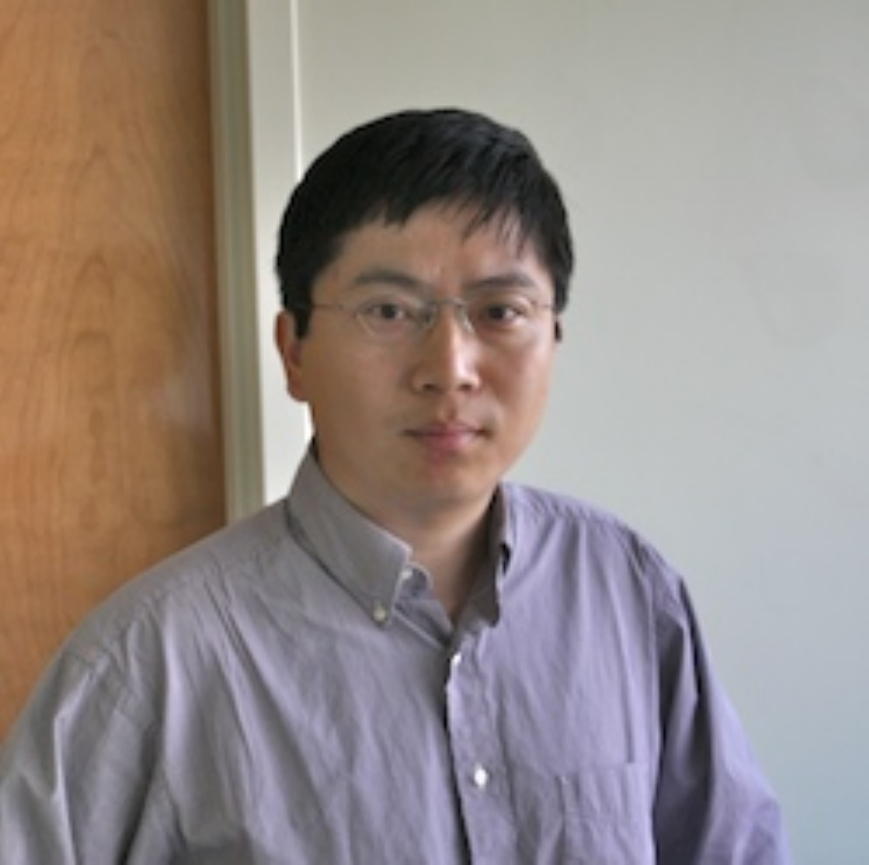}}]{Ying Nian Wu}
received a Ph.D. degree from the Harvard University in 1996. He was an Assistant Professor at the University of Michigan between 1997 and 1999 and an Assistant Professor at the University of California, Los Angeles between 1999 and 2001. He became an Associate Professor at the University of California, Los Angeles in 2001. From 2006 to now, he is a professor at the University of California, Los Angeles. His research interests include statistics, machine learning, and computer vision.
\end{IEEEbiography}

\vspace{-30pt}
\begin{IEEEbiography}[{\includegraphics[width=1in,height=1.25in,clip,keepaspectratio]{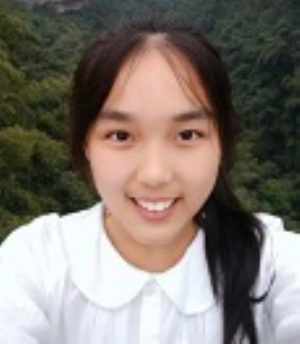}}]{Huilin Zhou}
received a B.S. degree in mathematics from the University of Electronic Science and Technology of China in 2019. Now, she is a Ph.D. candidate at the Shanghai Jiao Tong University. Her research interests include computer vision and machine learning.
\end{IEEEbiography}

\vspace{-30pt}
\begin{IEEEbiography}[{\includegraphics[width=1in,height=1.25in,clip,keepaspectratio]{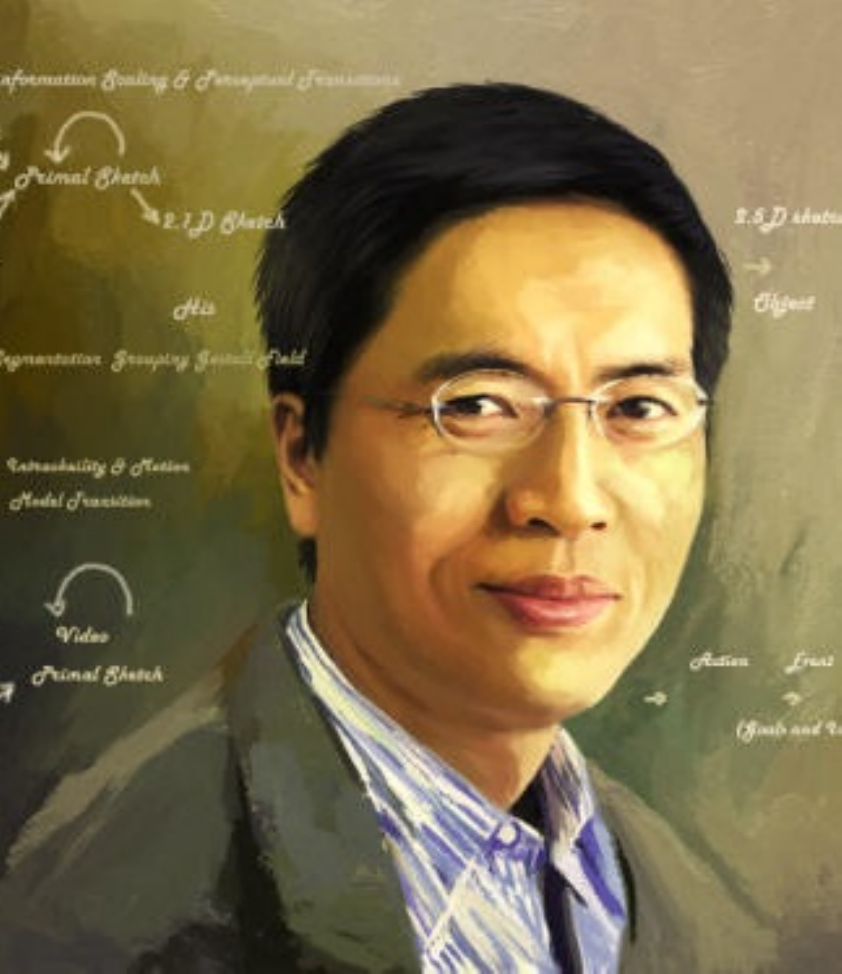}}]{Song-Chun Zhu} received a Ph.D. degree from Harvard University, and is a professor with the Department of Statistics and the Department of Computer Science at UCLA. His research interests include computer vision, statistical modeling and learning, cognition and AI, and visual arts. He received a number of honors, including the Marr Prize in 2003 with Z. Tu et. al. on image parsing,the Aggarwal prize from the Int'l Association of Pattern Recognition in 2008, twice Marr Prize honorary nominations in 1999 for texture modeling and 2007 for object modeling with Y.N. Wu et al., a Sloan Fellowship in 2001, the US NSF Career Award in 2001, and the US ONR Young Investigator Award in 2001. He is a Fellow of IEEE.
\end{IEEEbiography}
\vfill

\end{document}